\documentclass{article}

\PassOptionsToPackage{numbers,compress}{natbib}
\usepackage[preprint]{neurips_2026}

\usepackage[utf8]{inputenc} 
\usepackage[T1]{fontenc}    
\usepackage{hyperref}       
\usepackage{url}            
\usepackage{booktabs}       
\usepackage{amsfonts}       
\usepackage{nicefrac}       
\usepackage{microtype}      
\usepackage{xcolor}         
\usepackage{amsmath}
\usepackage{hyperref}
\usepackage{url}
\usepackage{amsthm}
\usepackage{wrapfig}  
\usepackage{booktabs,tabularx}
\usepackage{paralist}
\usepackage{graphicx}
\usepackage{array}  
\usepackage{changepage}
\usepackage{stfloats}
\usepackage{enumitem}
\usepackage{anyfontsize}
\usepackage{amssymb}
\usepackage{multirow}
\usepackage{listings}
\usepackage{pifont}
\usepackage[dvipsnames]{xcolor}
\usepackage{graphicx}
\usepackage{algorithm}
\usepackage{algorithmic}
\usepackage{listings}
\usepackage{booktabs}
\usepackage{graphicx}
\usepackage[table,xcdraw]{xcolor}
\usepackage{tcolorbox}
\tcbuselibrary{breakable,skins} 
\newcommand{\cmark}{\textcolor{ForestGreen}{\ding{51}}}
\newcommand{\xmark}{\textcolor{red!70!black}{\ding{55}}}
\newcommand{\evalpill}[2]{%
  \tcbox[on line, arc=2.5pt, boxrule=0.5pt, boxsep=0pt,
         left=3pt, right=3pt, top=0.4pt, bottom=0.4pt,
         colback=#1!12, colframe=#1!55, coltext=#1!75!black,
         fontupper=\scriptsize\bfseries]{#2}}
\newcommand{\evalAcc}{\evalpill{RoyalBlue}{Acc}}
\newcommand{\evalCost}{\evalpill{BlueGreen}{Cost}}
\newcommand{\evalStore}{\evalpill{Plum}{Storage}}
\newcommand{\samethanks}{\footnotemark[1]}
\title{GroupMemBench: Benchmarking LLM Agent Memory in Multi-Party Conversations}

\author{%
  Jingbo Yang\thanks{Equal Contribution. Correspondence to: \texttt{jingbo@ucsb.edu, henrylai@microsoft.com}.} \\
  UC Santa Barbara
  \And
  Kwei-Herng Lai\samethanks \\
  Microsoft
  \And
  Xiaowen Wang \\
  Microsoft
  \And
  Shiyu Chang \\
  UC Santa Barbara
  \AND
  Yaar Harari \\
  Microsoft
  \And
  Evgeniy Gabrilovich \\
  Microsoft
}

\begin{document}

\maketitle

\begin{abstract}
Large Language Model (LLM) agents increasingly serve as personal assistants and workplace collaborators, where their utility depends on memory systems that extract, retrieve, and apply information across long-running conversations. However, both existing memory systems and benchmarks are built around the dyadic, single-user setup, even though real deployments routinely span groups and channels with multiple users interacting with the agent and with each other. This mismatch leaves three properties of group memory unmeasured: (i)~group dynamics that go beyond concatenated one-on-one chats, (ii)~speaker-grounded belief tracking, where the per-user memory modeling is needed, and (iii)~audience-adapted language, where Theory-of-Mind shifts produce role-specific vocabulary. We introduce \textbf{GroupMemBench}, a benchmark that exposes all three. A graph-grounded synthesis pipeline produces multi-party conversations with controllable reply structure and conditions each message on per-user personas and target audiences. An adversarial query pipeline then binds every question to a specific asker across six categories, spanning multi-hop reasoning, knowledge update, term ambiguity, user-implicit reasoning, temporal reasoning, and abstention, and iteratively searches challenging, realistic queries that reflect comprehensive memory capability. Benchmarking leading memory systems exposes a sharp collapse: the strongest one reaches only 46.0\% average accuracy, with knowledge update at 27.1\% and term ambiguity at 37.7\%, while a simple \textsc{BM25} baseline matches or exceeds most agent memory systems. This indicates current memory ingestion erase the structural and lexical features group memory depends on, leaving multi-user memory far from solved.
\end{abstract}

\begin{center}
\href{https://huggingface.co/datasets/kimperyang/GroupMemBench}{\raisebox{-0.2\height}{\includegraphics[height=1em]{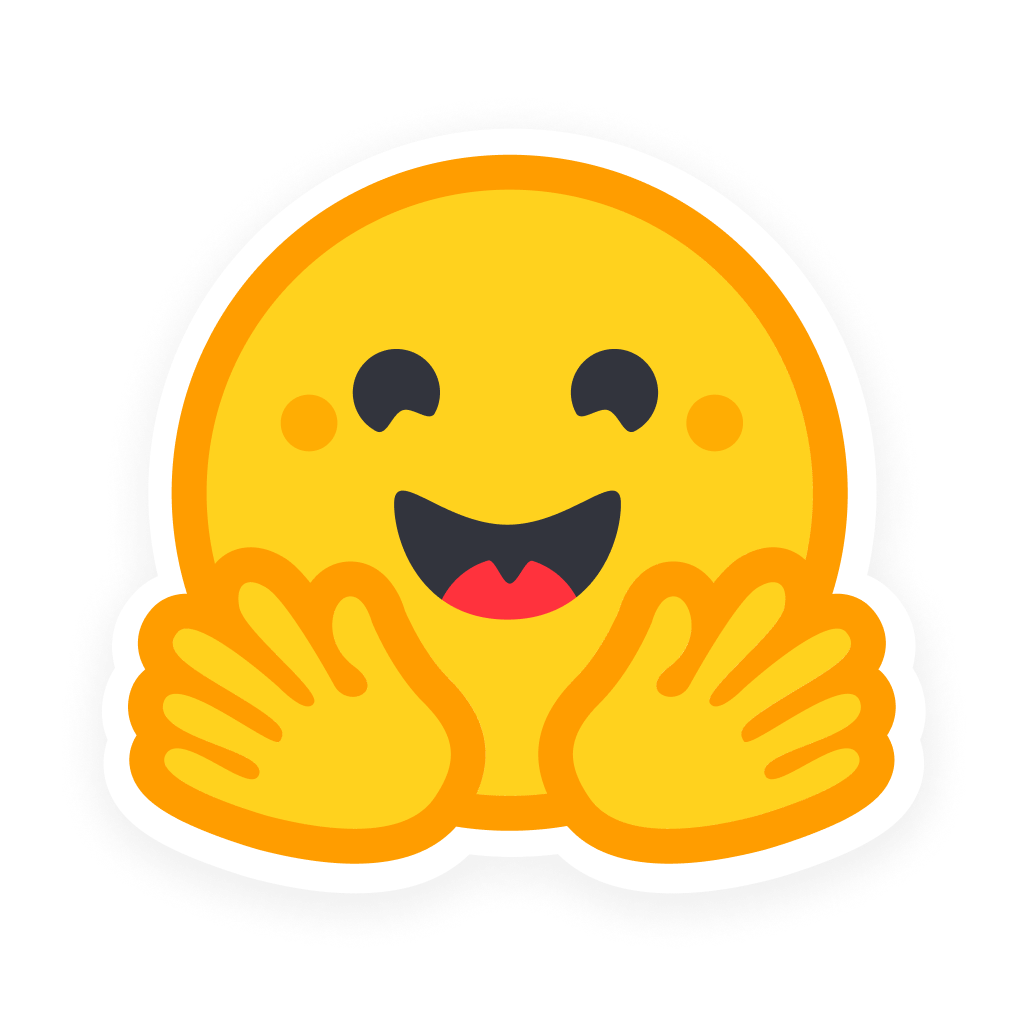}}~\textbf{Dataset}}
\quad
\href{https://github.com/UCSB-NLP-Chang/GroupMemBench}{\raisebox{-0.15\height}{\includegraphics[height=1em]{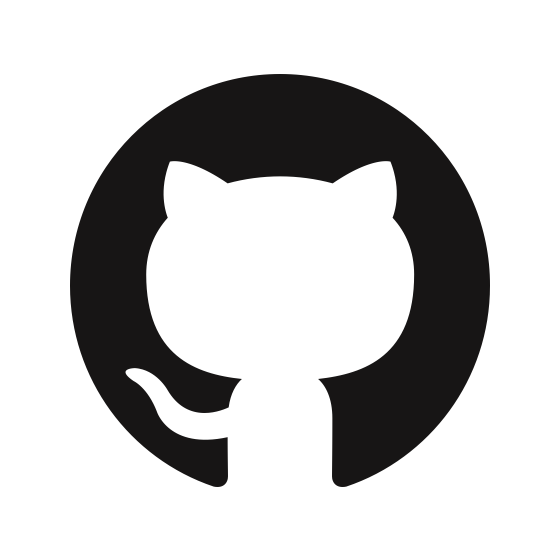}}~\textbf{Code}}
\end{center}

\section{Introduction}

Large Language Model (LLM)-powered agents have demonstrated strong 
capabilities across diverse tasks, both as personal 
assistants~\citep{wang2026openclaw,xu2026agent,ng2025trust,li2024personal,wang2024survey} and as 
workplace collaborators~\citep{stratton2024introduction}. To handle the 
conversational history that accumulates over long-horizon use, modern 
agents rely on memory mechanisms for extracting, consolidating, and 
retrieving information from past interactions. In many realistic 
scenarios, however, agents are deployed in group communication 
contexts (\emph{e.g.}, channels, threads, and project spaces), where multiple users 
interact with the same agent and with each other over time.
Despite this, current agent memory systems~\citep{huang2026rethinking,
latimer2025hindsight,chhikara2025mem0} and their 
evaluations~\citep{he2026memoryarena,zhao2026ama,ai2025memorybench,
wu2024longmemeval} almost exclusively target the dyadic, single-user 
setting, leaving the multi-party memory dynamics largely unmeasured.

The transition from single-user memory to multi-party group memory 
introduces qualitatively new complexities along three dimensions. 
\ding{182}~\textbf{Group dynamics.} Single-user interactions typically 
follow a straightforward instruction, i.e., response paradigm, whereas 
multi-party dialogues encompass richer social patterns such as open 
discussions, threaded debates, and collaborative consensus-building. 
The collective task flow therefore cannot be reduced to a flat sequence 
of turns or a set of parallel one-on-one streams. 
\ding{183}~\textbf{Speaker-grounded belief tracking.} The identity of 
the message author becomes part of its content: identical statements 
from different speakers may warrant entirely different memory updates, 
and identical queries from different askers may warrant different 
answers. An agent must therefore contextualize its reasoning around the 
specific user's preferences, role, and historical knowledge, rather 
than treating the conversation as an unattributed text stream. 
\ding{184}~\textbf{Audience-adapted language.} Human speakers naturally 
exhibit Theory of Mind (ToM)~\citep{frith2005theory,kim2023fantom,shinoda2025tomato,clark1991grounding}, adapting their 
wording to assumed shared knowledge with their audience. In workplace 
settings this surfaces as peer-specific jargon. For example, "token" 
may denote a unit of text in NLP discussions and an authentication 
credential in system-design discussions. This produces systematic lexical 
shifts across roles that pose a direct challenge to retrieval-augmented 
generation~\citep{lewis2020retrieval,gutierrez2024hipporag}.

\begin{figure}[t]
    \centering
    \includegraphics[width=\textwidth]{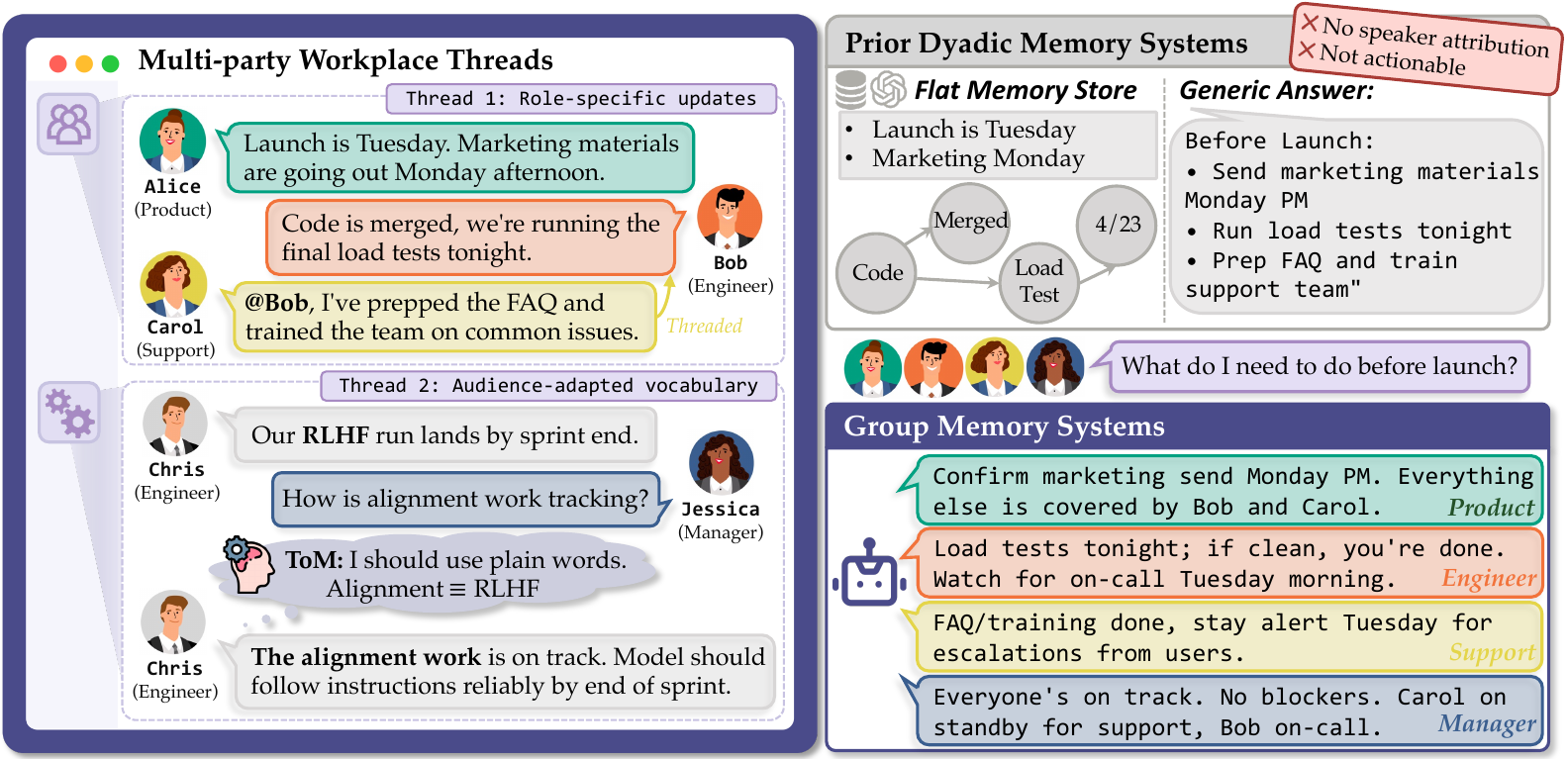}
    \vspace{-15pt}
    \caption{Dyadic memory systems are inadequate for group memory, which demands joint modeling of the collective task flow and per-user memory states. The multi-party attributes, such as Theory-of-Mind, also make this fundamentally harder than a multi-user extension of dyadic memory.}
    \label{fig:teaser}
\end{figure}

To address the complexities, we benchmark existing memory mechanisms in multi-party group settings for a central question: \emph{can current memory systems serve the right user, with the right belief, under the right interpretation, when memory is built from group conversations?} We decompose this into three sub-questions, each paired with a benchmark mechanism designed to isolate the corresponding source of difficulty.
\textbf{(Q1)~Can memory systems capture group dynamics that go beyond concatenated one-on-one chats?} We generate conversations with a controllable reply structure that induces threaded debates, multi-user replies, and cross-topic shifts, rather than a flat turn sequence.
\textbf{(Q2)~Can memory systems condition extraction, consolidation, and retrieval on speaker identity?} We bind each query to a specific asking user and introduce three speaker-aware adversarial categories: \emph{user-implicit reasoning} hides the asker behind first-person references, \emph{knowledge update} maintains incompatible preferences across users, and \emph{term ambiguity} makes a term's referent depend on who is asking.
\textbf{(Q3)~Can memory systems remain robust to ToM-driven lexical shifts across roles?} We condition messages on per-user personas and explicit target audiences so that the same underlying fact appears in role-specific wording; \emph{term ambiguity} then asks questions in the asker's vocabulary while the supporting evidence is recorded in another role's phrasing, creating a controlled retrieval stress test.
It is worth mentioning that benchmark difficulty is enforced rather than asserted: each accepted query must defeat a competent retrieval baseline through a Solve-Judge-Refine loop (Section~\ref{subsubsec:shared_pipeline}).

Together these design choices yield \textbf{GroupMemBench} 
(Figure~\ref{fig:teaser}). Benchmarking leading memory systems on it reveals 
systematic gaps in group memory: even the best-performing system achieves 
only 46.0\% average accuracy. Abstention is largely preserved (above $80\%$),
but content-bearing categories such as \emph{knowledge update} and
\emph{term ambiguity} drop to 27.1\% and 37.7\%. We expect progress on GroupMemBench 
to reflect improved speaker-conditioned retrieval, belief tracking across users, 
and structure-aware retrieval in multi-party conversations.

To summarize, we make the following contributions in this paper:
\begin{itemize}[leftmargin=10pt,topsep=2pt,itemsep=1pt]
     \item We define group-memory evaluation along three axes: dynamics, speaker grounding, audience adaptation.
  \item We introduce \textbf{GroupMemBench}: multi-party conversations with user-bound queries.
  \item We construct the benchmark via reply-structure control, speaker-aware categories, and persona/audience conditioning.
  \item We enforce difficulty with Solve-Judge-Refine and benchmark leading memory systems.
\end{itemize}

\begin{table}[h]
\caption{Comparison of GroupMemBench with representative long-term memory benchmarks.
\textbf{Multi-User}: dialogues feature multiple human-modeled speakers.
\textbf{ToM Adaptation}: speakers adapt vocabulary and explanations based on the roles and prior knowledge.
\textbf{Threaded}: messages form a post-and-reply hierarchy rather than a flat turn-by-turn timeline.
\textbf{User-Cond. Q.}: the gold answer can change with the querying user's identity.
\textbf{\#\,Topics}: number of distinct conversation topics organized in the corpus or sub-task types.
\textbf{\#\,Turns}: total messages/utterances in the corpus; ``---'' indicates the benchmark does not report a turn-level statistic.
\textbf{Avg. Length}: average tokens per dialogue (values taken from~\citep{zhao2026ama}).
\textbf{Eval. Dimensions}: \evalAcc{} = answer accuracy; \evalCost{} = LLM and embedding inference cost incurred while constructing memory; \evalStore{} = processed memory overhead.}
\label{tab:bench_compare}
\centering
\footnotesize
\setlength{\tabcolsep}{4pt}
\renewcommand{\arraystretch}{1.35}
\resizebox{\linewidth}{!}{%
\begin{tabular}{l c c c c c c c c}
\toprule
\textbf{Benchmark} & \textbf{Multi-User} & \textbf{ToM Adaptation} & \textbf{Threaded} & \textbf{User-Cond. Q.} & \textbf{\#\,Topics} & \textbf{\#\,Turns} & \textbf{Avg. Length} & \textbf{Eval. Dimensions} \\
\midrule
LoCoMo~\citep{maharana2024evaluating}        & \xmark & \xmark & \xmark & \xmark & 1            & 15K           & 9K            & \evalAcc \\
LongMemEval~\citep{wu2024longmemeval}        & \xmark & \xmark & \xmark & \xmark & 6            & 11K           & 115K          & \evalAcc \\
MemoryAgentBench~\citep{hu2025evaluating}    & \xmark & \xmark & \xmark & \xmark & 4            & ---           & 100K--500K    & \evalAcc~\evalCost \\
MemoryBench~\citep{ai2025memorybench}        & \xmark & \xmark & \xmark & \xmark & 3            & ---           & 30K--380K     & \evalAcc~\evalCost \\
EverMemBench~\citep{hu2026evermembench}      & \cmark & \xmark & \xmark & \xmark & 15           & 51K           & 845K          & \evalAcc \\
\midrule
\rowcolor{blue!8}
\textbf{GroupMemBench (Ours)}                & \cmark & \cmark & \cmark & \cmark & \textbf{123} & \textbf{120K} & \textbf{2.7M} & \evalAcc~\evalCost~\evalStore \\
\bottomrule
\end{tabular}%
}
\end{table}

\section{Related Work}

\paragraph{Agent Memory Systems.}
Agent memory systems persist useful facts across long conversations so that an LLM agent can answer later queries without re-processing the full history~\citep{zhong2024memorybank,zhang2025survey,tan2025prospect,jiang2025personamem,jiang2026magma}. Mainstream designs cast ingestion as online fact extraction over a structured store: MemGPT~\citep{packer2023memgpt} manages a tiered working/archival memory with explicit paging, Mem0~\citep{chhikara2025mem0} extracts atomic facts and applies add/update/delete operations during ingestion, A-Mem~\citep{xu2025mem} links extracted notes into a self-organizing graph, HippoRAG~\citep{gutierrez2024hipporag} grounds dense retrieval in an entity-centric knowledge graph for multi-hop recall, and Hindsight~\citep{latimer2025hindsight} adds a reflection stage that rewrites consolidated memory over time. A recent position paper~\citep{huang2026rethinking} surveys this landscape and argues for richer procedural memory. All of these systems were designed around dyadic interactions and inherit two structural assumptions that break in group settings: \texttt{user\_id} is used as an isolated namespace that separates data stores rather than as a semantic attribute that conditions extraction and retrieval, and conflicting statements are resolved by overwriting older entries rather than preserving coexisting beliefs from different speakers. GroupMemBench specifically probes these two assumptions by requiring systems to attribute each fact to its speaker and to keep concurrent preferences addressable by the querying user.

\paragraph{Long-term Memory Benchmarks.}
Long-term memory benchmarks have scaled rapidly along context length, session count, and reasoning depth while keeping the interaction format strictly dyadic. LoCoMo~\citep{maharana2024evaluating} and LongMemEval~\citep{wu2024longmemeval} expand conversations to hundreds of sessions and test multi-session recall; MemBench~\citep{tan2025membench}, MemoryAgentBench~\citep{hu2025evaluating}, MemoryArena~\citep{he2026memoryarena}, AMA-Bench~\citep{zhao2026ama}, and MemoryBench~\citep{ai2025memorybench} broaden the question taxonomy to include multi-hop reasoning, temporal alignment, knowledge updates, and continual learning. Across this line of work, however, each utterance has a single author and every query is issued by ``the user'': there is no notion of speaker-conditioned retrieval, coexisting preferences, or role-dependent vocabulary, so memory systems tuned on these benchmarks are never forced to model \emph{who} said what to \emph{whom}. EverMemBench~\citep{hu2026evermembench} is the closest prior effort to evaluate memory under multi-party dialogue, yet its speakers are homogeneous chat participants without persona or Theory-of-Mind modeling, leaving agents untested on user-implicit reasoning and on vocabulary that shifts with the addressee. GroupMemBench fills this gap by pairing graph-grounded multi-user synthesis with adversarial queries that explicitly target speaker-dependent answers.

\section{GroupMemBench}

\subsection{Problem Formulation}
\label{sec:problem_formulation}


\paragraph{Group Memory Environment}
\label{subsec:group_memory_environment}
Let $\mathcal{U} = \{u_1, \dots, u_K\}$ denote the set of $K$ users in a group, and let $u_a$ denote the memory-augmented agent. The conversation is a chronologically ordered sequence $\mathcal{C} = (m_1, \dots, m_N)$, where each message is a tuple $m_i = (a_i, t_i, c_i)$ with author $a_i \in \mathcal{U} \cup \{u_a\}$, timestamp $t_i$, and textual content $c_i$. Each $m_i$ may correspond to either a user--user interaction or a user--agent interaction, and every message, including those authored by $u_a$, enters the shared history and conditions all subsequent turns, so the agent's replies evolve as a Markov decision process over $\mathcal{C}$. This contrasts with the dyadic setting ($K=1$, \emph{i.e.}, one user plus the agent), where $a_i$ trivially alternates and the author tag carries no information.

Starting from an empty state $\mathcal{M}_0$, the agent maintains a memory $\mathcal{M}_i$ that is incrementally updated after each message: $\mathcal{M}_i = f_{\text{update}}(\mathcal{M}_{i-1}, m_i)$.
The update function $f_{\text{update}}$ is left abstract, as concrete implementations vary across memory systems (\emph{e.g.}, fact extraction, graph updates, or raw append).

\paragraph{Task Definition}
\label{subsec:task_definition}
We evaluate the memory state through memory-grounded question answering. After observing a prefix $\mathcal{C}_{1:T}$ and forming memory $\mathcal{M}_T$, the agent receives a query $q$ issued by a specific user $u_q \in \mathcal{U}$ (with $q \in \{1,\dots,K\}$), and produces an answer
$a = f_{\text{ans}}(q, u_q, \mathcal{M}_T)$,
which is itself appended to $\mathcal{C}$ as the next message and thereby updates $\mathcal{M}_{T+1}$ and shapes the future trajectory of the conversation. The explicit dependence on $u_q$ is the key departure from dyadic formulations, where $u_q$ is a constant and is typically dropped. In the group setting, the ground-truth answer may differ across querying users even for the same $q$: resolving references in $q$, interpreting vocabulary, and ranking retrieved facts all depend on $u_q$'s identity, role, and prior history. A correct system must therefore treat $u_q$ as a first-class input, not an implicit context.

\begin{figure}[t]
    \centering
    \includegraphics[width=0.97\textwidth]{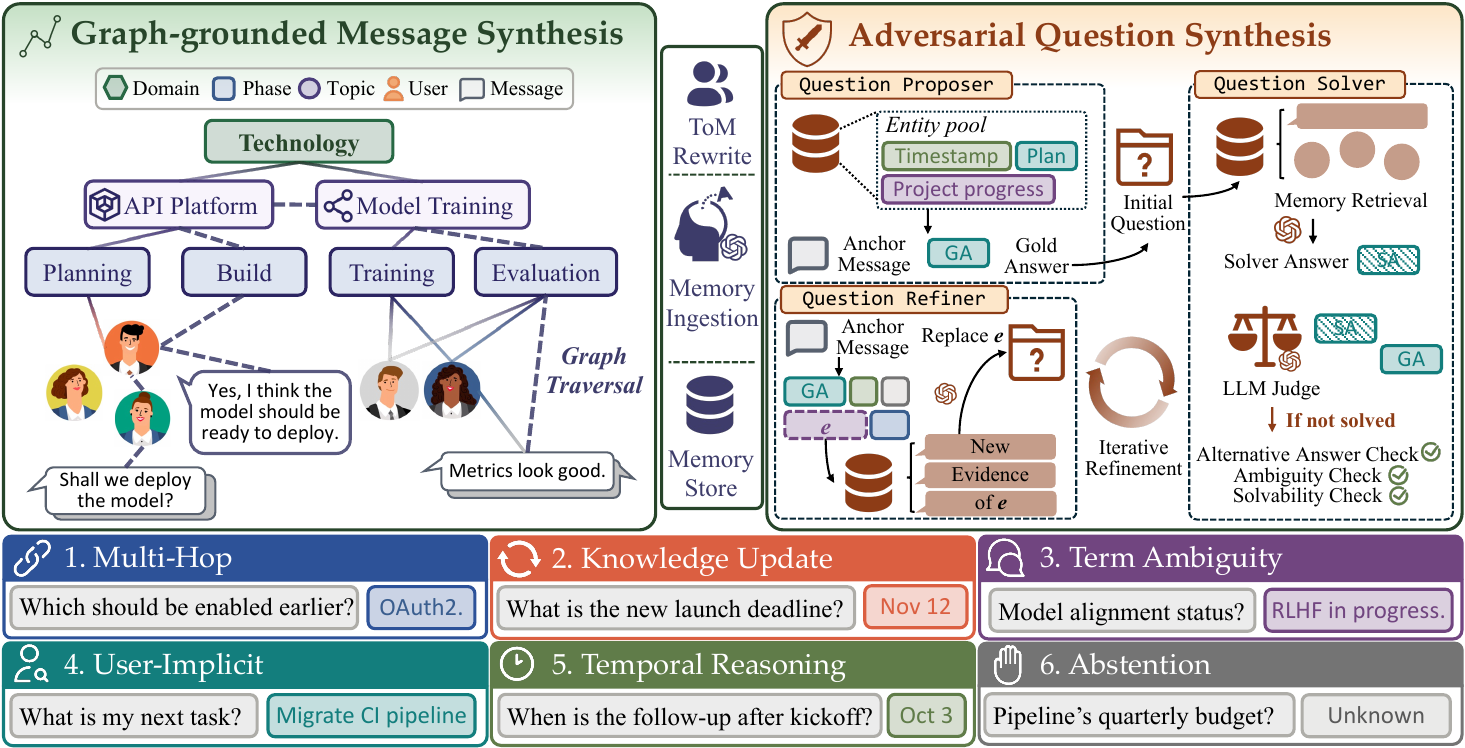}
    \vspace{-5pt}
    \caption{Overview of the GroupMemBench data synthesis pipeline .}
    \label{fig:method}
\end{figure}

\subsection{Synthesizing Multi-Party Conversations}
\label{sec:dataset_construction}
We generate multi-party conversations by separating structure from text. A typed graph encodes domains, topics, phases, and users, and is augmented during generation with message nodes connected by reply-to and authored-by links. Each new message begins with a graph traversal that selects the author, recipients, and thread/phase context; a language model then realizes the message conditioned on the retrieved situational history. This factorization makes group-level phenomena controllable while maintaining long-horizon coherence.

\paragraph{Knowledge Graph Representation.}
We represent the conversational environment as a directed graph
$\mathcal{G} = (\mathcal{V}, \mathcal{E})$ that grows over time. The structural
backbone $\mathcal{V}_0$ contains four node types:\emph{domains} ($d$),
\emph{topics} ($\tau$), \emph{phases} ($\phi$), and \emph{users}
($u$). The graph $\mathcal{G}$ is augmented during generation with a fifth type, \emph{messages}($m$), as they are produced. Each phase node carries an owner, a status drawn
from a category-specific schema (\textsc{Risks}, \textsc{Milestone}, or
\textsc{Work Items}, each with its own status progression), a target date, and
a temporal window over which messages can be timestamped. Each user node
carries a contextual profile $\mathcal{C}_u$ that records persona attributes
(role, tone, style, expertise) and prior message history; tone and style
combinations are drawn without replacement so every user has a distinct voice.
The edge set $\mathcal{E}$ encodes hierarchical relations
($d \!\xrightarrow{\text{has\_topic}}\! \tau$,
$\tau \!\xrightarrow{\text{has\_phase}}\! \phi$),
membership relations
($u \!\leftrightarrow\! d$, $u \!\leftrightarrow\! \phi$),
cross-domain bridges via shared categories
($d \!\xrightarrow{\text{related\_to}}\! d'$),
and, once messages exist, threading and authorship relations
($m \!\xrightarrow{\text{reply\_to}}\! m'$,
$m \!\xrightarrow{\text{authored\_by}}\! u$). Grounding generation in
$\mathcal{G}$ allows messages to reason over user-specific trajectories, phase
progress, and inter-project dependencies rather than being conditioned on
isolated prompts. The full schema is given in
Table~\ref{tab:relations}, Appendix~\ref{appendix:graph_schema}.

\paragraph{Path-Based Context Sampling.}
Each new message is generated by sampling a traversal path over $\mathcal{G}$
that determines the message's author, recipients, and conversational
situation. We define three path templates corresponding to the primary
dynamics of group dialogue, and sample among them with mixture weights
$(0.2,\,0.6,\,0.2)$:
\begin{itemize}[leftmargin=10pt, itemsep=1pt]
    \item \textbf{Channel post}
    ($d \!\rightarrow\! \tau \!\rightarrow\! \phi \!\rightsquigarrow\! u$):
    opens a new sub-thread by walking from a domain through one of its topics
    to an active phase, biased by a \emph{post-need} weight that favours
    phases with few or no posts, and then sampling an author from the phase's
    active members.
    \item \textbf{Threaded reply}
    ($d \!\rightarrow\! \tau \!\rightarrow\! \phi \!\rightsquigarrow\! m \!\rightarrow\! u$):
    walks to a phase under a \emph{phase-heat} weight (favouring phases with
    recent activity and unanswered posts), then selects a parent message $m$
    from the phase's history via preferential attachment combined with
    linear recency decay and a post-over-reply bias, and finally chooses a
    responder distinct from the parent author and the last few speakers.
    \item \textbf{Cross-project bridge}
    ($u \!\rightarrow\! \phi \!\rightarrow\! \tau \!\rightarrow\! d \!\xrightarrow{\text{rel.}}\! d' \!\rightarrow\! \tau' \!\rightarrow\! \phi' \!\rightarrow\! u'$):
    anchors at a user $u$, follows the \texttt{related\_to} edge induced by
    shared domain categories to a sibling project $d'$, and lands on a target
    user $u'$ with the same role as $u$ in a temporally consistent phase
    $\phi'$, capturing cross-team continuity such as role-to-role
    consultations.
\end{itemize}
A sampled path yields a set of contextual nodes that jointly define the
generation condition. From these nodes we retrieve the corresponding metadata
and the most recent messages within a fixed-size context window, providing the
local turn-by-turn history. To prevent the long, narrow phases from devolving
into paraphrastic loops, we additionally maintain a per-phase memory:
at first use of a phase, we prompt the LLM once to enumerate a small budget of
distinct sub-issues that a real team would discuss, and every few messages we
refresh a short rolling summary of which points are already \textsc{Decided},
which questions remain \textsc{Open}, and which commitments are still
\textsc{Pending}. This summary, together with the active sub-issue, is
appended to the retrieved history to form the situational context
$\mathcal{S}_u$ used at the current generation step. Sampling probabilities,
weight functions, and graph-walk details are provided in
Appendix~\ref{appendix:path_sampling}.

\paragraph{Message Generation.}
Given the situational context $\mathcal{S}_u$, the user profile
$\mathcal{C}_u$, and the path type, we construct a prompt that encodes the
author's persona, the current topic and phase, the phase's stage
(\emph{early}, \emph{middle}, or \emph{late}, derived from a Poisson-bursty
timestamp schedule) and status goal, the retrieved context, and the intended
conversational scenario (full templates in Appendix~\ref{appendix:prompt}).
The prompt is passed to a language model (\textsc{GPT-5} in our
implementation) to sample a candidate message
\begin{equation}
\hat{m} \sim p\bigl(\,\cdot \mid \mathrm{prompt}(\mathcal{S}_u,\,\mathcal{C}_u,\,\mathrm{stage},\,\mathrm{path})\bigr).
\end{equation}
For replies, the prompt explicitly references the parent message to enforce
threading coherence. The accepted message is instantiated as a new node in
$\mathcal{G}$, annotated with author, timestamp, and path type, and linked to
the relevant users, phase, and parent message. To keep the corpus from being
artificially clean, the policy layer stochastically injects four classes of
realism perturbations: (i) \emph{noise messages} ($\sim$5\%) that contain a
subtle factual mistake, stale reference, or off-topic drift; (ii)
\emph{disagreement turns} ($\sim$25\% of replies) that push back on or
conditionally accept a prior point; (iii) \emph{stylistic imperfections}
($\sim$20\%) such as typos, fragments, or self-corrections, alongside
ultra-short replies and tone jitter; and (iv) \emph{decision changes}
($\sim$5\% of eligible phases) in which a previously committed decision is
deliberately reversed mid-phase, with the reversal recorded as ground-truth
metadata for downstream evaluation.
Using this pipeline, we synthesize group conversations spanning four domains (Technology, Healthcare, Manufacturing, Finance), with each message labeled as either an initiating \emph{post} or a threaded \emph{reply}. Per-domain corpus statistics are reported in Appendix~\ref{appendix:dataset_stats}.

\begin{wrapfigure}{r}{0.45\linewidth}
\vspace{-1.2em}
\centering
\includegraphics[width=\linewidth]{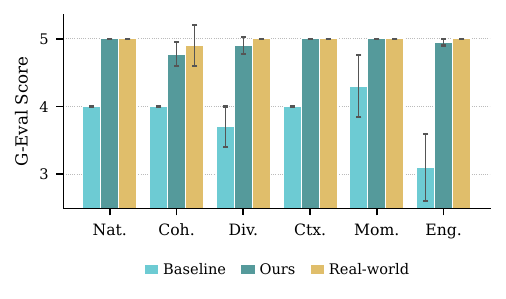}
\vspace{-20pt}
\caption{G-Eval scores across six dimensions. Our graph-guided synthesis (four domains) closely tracks the real-world upper bound and substantially outperforms the single-prompt baseline. Scores averaged over 10 seeds; shaded bands indicate $\pm 1$ std.}
\vspace{-1.2em}
\label{fig:geval}
\end{wrapfigure}
\paragraph{Quality Assessment.}
We adapt G-Eval~\citep{liu2023g} to the group-chat setting and assess synthesized dialogues along six dimensions chosen to reflect properties specific to multi-party interaction: \emph{Naturalness}, \emph{Coherence}, \emph{Diversity}, \emph{Contextual Relevance}, \emph{Momentum}, and \emph{Engagingness}, each scored from 0 to 5 by GPT-5 as judge (rubric in Appendix~\ref{appendix:geval}). We compare our four synthesized domains against two references: a \emph{Single-Prompt Baseline}, which generates group messages from a single monolithic prompt without graph grounding, and a \emph{Real-World Upper Bound} drawn from authentic group chat logs of our own in 1 month horizon. Each score is averaged over 10 seeds at fixed temperature. As shown in Figure~\ref{fig:geval}, our synthesis matches the real-world conversations on most dimensions and substantially outperforms the single-prompt baseline, confirming that graph-guided generation yields conversations that are structurally and stylistically realistic.

\subsection{Adversarial Query Construction}
\label{subsec:query_construction}

To systematically evaluate distinct failure modes in memory-augmented agents, we synthesize queries across six orthogonal dimensions. We employ a shared adversarial pipeline that decouples benchmark difficulty from human labeling effort: a query is only accepted into the benchmark if a competent retrieval-based baseline empirically fails to answer it.

\subsubsection{Shared Generation Pipeline}
\label{subsubsec:shared_pipeline}
Given a multi-party conversation, each query is generated through a rigorous four-stage process:
\ding{182} \textbf{Anchor Selection:} A lightweight filter identifies candidate messages containing the target entities, which an LLM classifier then verifies for concrete answerability.
\ding{183} \textbf{Initial Composition:} A type-specific question proposer LLM drafts a natural language question, based on the anchor message as the initial evidence and target entity as the gold answer.
\ding{184} \textbf{Solve--Judge--Refine Loop:} A baseline agent attempts the question using retrieved context. An LLM judge evaluates the response. If the solver succeeds, the query is iteratively refined to increase difficulty. If it fails, the query is accepted.
\ding{185} \textbf{Output:} For each accepted query, we retain the final question, gold answer, querying user ID, and a per-round evidence trace.

\subsubsection{Dimensions and Type-Specific Strategies}
\label{subsubsec:dimensions_strategies}
Within this harness, we instantiate six query types, each targeting a distinct capability of group memory; type-specific generation and validation details are deferred to Appendix~\ref{appendix:prompt}.

\begin{itemize}[leftmargin=10pt, itemsep=1pt]
    \item \textbf{Multi-Hop Reasoning:} Tests the agent's ability to chain evidence scattered across distant messages and threads, where the answer is reachable only by linking a target entity to a separate ``pivot'' message that refers to it indirectly.

    \item \textbf{Knowledge Update:} Tests whether the agent tracks dynamic state changes and prefers the latest consensus over outdated facts. Each query is anchored to a documented decision change (\emph{e.g.}, a rescheduled deadline) and may probe the current state, the trigger, or the prior value.

    \item \textbf{Term Ambiguity:} Tests robustness to speaker-conditioned vocabulary. The question is phrased in the asker's role-specific terminology while the supporting evidence uses a different role's variant (\emph{e.g.}, engineer's ``PPO'' vs.\ manager's ``alignment''), so the agent must resolve referents through understanding and linking the memories from different speakers, not the surface form.

    \item \textbf{User-Implicit Reasoning:} Tests whether the agent can resolve first-person references to a specific user. Queries are cast in the first person, emphasizing memory modeling of long-term user profile.

    \item \textbf{Temporal Reasoning:} Tests chronological reasoning over the dialogue. Questions are grounded on both relative time expressions (\emph{e.g.}, ``by EOD this Friday'') and timestamps attached to messages.

    \item \textbf{Abstention:} Tests resistance to hallucination at the edge of the agent's knowledge. The query targets a fact absent from the conversation, requiring the agent to refuse rather than fabricate.
\end{itemize}

\section{Experiments}

\subsection{Experimental Setup}
\label{subsec:exp_setup}

\paragraph{Baselines.}
We compare against two families of memory approaches that represent the dominant paradigms in the current literature. \ding{182}~\textbf{retrieval-augmented generation (RAG)} baselines, which treat the raw conversation as a passive document store and retrieve relevant chunks at query time. We include regular RAG with different retrievers, and graph RAG, which augments dense retrieval with an entity-centric knowledge graph. \ding{183}~\textbf{agent memory systems} that actively extract, consolidate, and index facts during ingestion: \emph{MemGPT}~\citep{packer2023memgpt}, \emph{Mem0}~\citep{chhikara2025mem0}, \emph{A-Mem}~\citep{xu2025mem}, etc.

\paragraph{Evaluation Metrics.}
\textbf{(1) Accuracy:} question-answering accuracy judged by GPT-5 against reference answers~\citep{liu2023g}, reported per category and averaged. We validate the LLM judge reliability on a human-annotated subset (Appendix~\ref{appendix:llm_judge}).
\textbf{(2) Efficiency:} \emph{ingestion cost} (USD) sums all LLM and embedding calls during memory construction, with query-time calls excluded; \emph{storage footprint} (GB) is the on-disk size of each baseline's per-domain store. Token-to-USD pricing, the proxy used for token accounting, and the procedure for apportioning shared databases are detailed in Appendix~\ref{appendix:efficiency_table}.

\paragraph{Implementation Details.}
To isolate methodological differences from backbone differences, we standardize the underlying models across all baselines that do not ship with pretrained checkpoints. During \emph{memory ingestion}—including fact extraction, summarization, and graph construction—all systems use \texttt{gpt-4o-mini} as the backbone language model. At \emph{query time}, the answering agent uses \texttt{GPT-5}, and all dense retrieval components use OpenAI's \texttt{text-embedding-3-large}. All baselines are evaluated following the official implementations.


\definecolor{raghdr}{HTML}{FCE7DA}    
\definecolor{agenthdr}{HTML}{E4ECF8}  

\begin{table}[t]
\caption{Accuracy (\%) of memory systems on GroupMemBench across six query categories. Best per column in \textbf{bold}, second-best \underline{underlined}.}
\label{tab:main_results}
\centering
\scriptsize
\setlength{\tabcolsep}{4pt}
\renewcommand{\arraystretch}{1.25}
\resizebox{\linewidth}{!}{%
\begin{tabular}{l ccccccc}
\toprule
\textbf{Method} & \textbf{Multi-Hop} & \textbf{Update} & \textbf{Ambiguity} & \textbf{Implicit} & \textbf{Temporal} & \textbf{Abstention} & \textbf{Average} \\
\midrule
\rowcolor{raghdr} \multicolumn{8}{l}{\textit{RAG-based Methods}} \\
BM25                               & \underline{40.11} & \underline{25.23} & 14.15          & 40.82          & \textbf{54.94} & \underline{77.98} & \underline{43.22} \\
text-embedding-3-large        & 36.26          & 23.36          & 21.70          & \textbf{46.94} & \underline{32.72} & 75.23          & 38.04 \\
GraphRAG~\citep{edge2024local}                          & 12.09          & 14.02          & 19.81          & 14.29          & \phantom{0}5.56 & 66.97          & 20.56 \\
\midrule
\rowcolor{agenthdr} \multicolumn{8}{l}{\textit{Agent Memory Systems}} \\
Mem0~\citep{chhikara2025mem0}                           & 21.98          & \phantom{0}4.67 & 11.32         & 20.41          & 16.67          & \textbf{82.57} & 25.73 \\
MemGPT~\citep{packer2023memgpt}                         & 22.53          & 17.76          & 20.75          & 28.57          & 12.42          & \underline{77.98} & 28.15 \\
A-Mem~\citep{xu2025mem}                                 & 35.16          & 22.43          & 26.42          & \textbf{46.94} & 23.46          & 67.89          & 35.10 \\
HippoRAG~\citep{gutierrez2024hipporag}                  & 39.56          & \textbf{27.10} & \underline{30.19} & \underline{42.86} & 29.63          & 75.23          & 39.72 \\
Hindsight~\citep{latimer2025hindsight}                  & \textbf{42.31} & 17.76          & \textbf{37.74} & 40.82          & \textbf{54.94} & 77.06          & \textbf{46.01} \\
\bottomrule
\end{tabular}%
}
\end{table}

\subsection{Main Results}
\label{sec:results}
Table~\ref{tab:main_results} reports cross-domain accuracy across the six query categories (per-domain numbers in Appendix~\ref{appendix:per_domain}). We organize the discussion around the three questions raised in the Introduction: across baselines, current memory systems \textbf{(Q1)} flatten group structure during ingestion, \textbf{(Q2)} fail to condition extraction and retrieval on speaker identity, and \textbf{(Q3)} cannot resolve lexical shifts even after the gold message is retrieved. The two categories where systems perform best, Temporal and Abstention, are also the ones least dependent on multi-party structure: Temporal answers are pinned to a date, and Abstention only asks whether the answer is in memory at all.

\textbf{(Q1) Group structure is flattened during ingestion.}
No system exceeds 50\% on Multi-Hop, the category that most directly probes evidence chained across thread structure. The deeper signal is that BM25, with no semantic processing during ingestion, beats four of the five agent memory systems on average~\ref{app:information-loss}. Agent memory was designed to compress one user's history into self-contained facts, and that compression discards exactly the structure the multi-party setting needs.

\textbf{(Q2) Memory systems do not condition on the speaker.}
Three categories isolate this gap. On \emph{User-Implicit}, the best agent memory system matches a bare \texttt{text-embedding-3-large} retriever, indicating no real per-user modeling and the extra ingestion machinery contributes nothing here. On \emph{Knowledge Update}, where decisions usually shift through back-and-forth among multiple users, no system passes 28\%; Mem0 collapses to 4.67 because its extractor appends new statements alongside obsolete ones without speaker-conditioned consolidation. \emph{Term Ambiguity} reinforces the picture: dyadic memory has no notion of per-speaker referents, so the ambiguity caused by a word like ``token'' is irrecoverable. The two cross-category outliers tell the same story: Mem0's high Abstention is conservative storage rather than calibrated refusal, and Hindsight's gains on Multi-Hop and Ambiguity come from aggregating paraphrased notes across messages, the same aggregation that locks old decisions and drives its Update score below 18\%.

\textbf{(Q3) Lexical shifts are the residual failure even when retrieval succeeds.}
Term Ambiguity stays below 38\% across all baselines, and the gap persists even after factoring out retriever quality: as we show in Section~\ref{sec:failure_analysis}, $P(\mathrm{correct}\mid\mathrm{gold\ retrieved})$ remains uniformly low for this category, making it the only type where representation, not retrieval, is the wall. As these lexical shifts naturally attack the retrieval process, this further raises a future research direction on how to handle multi-party cognitive differences. We believe the solution to this should be a collaborative evolution of both memory modeling (\emph{e.g.}, agent memory ingestion) and agent reasoning improvement.

\subsection{Performance-Efficiency Trade-off}
\label{subsec:tradeoff}

\begin{figure}[t]
    \centering
    \includegraphics[width=\linewidth]{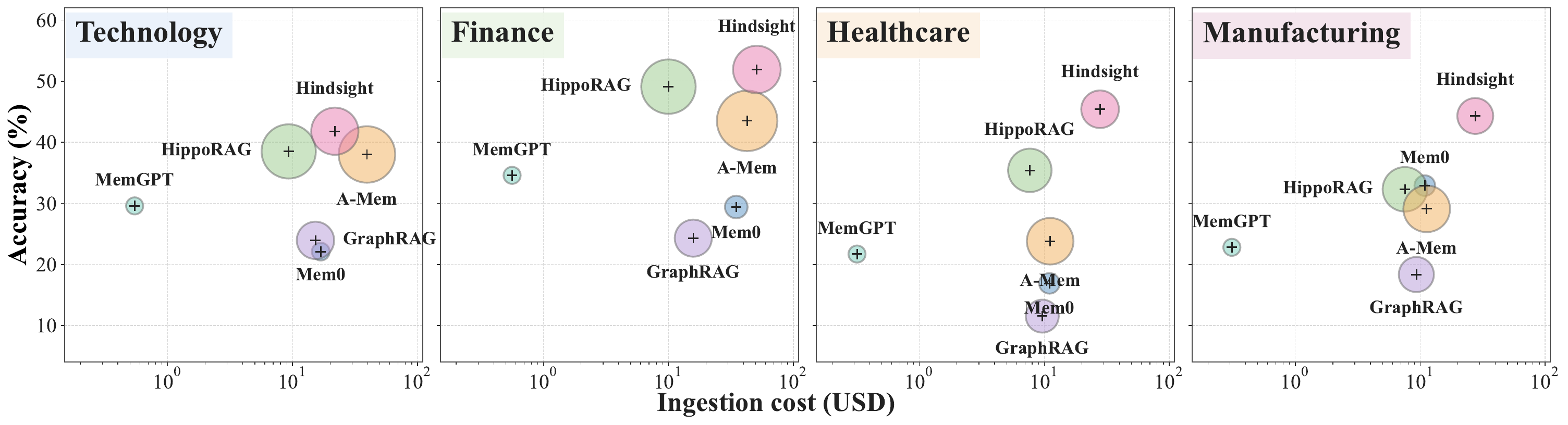}
    \vspace{-20pt}
    \caption{Performance--efficiency trade-off across the four domains. Each marker is one of six baselines (five agent memory systems plus \textsc{GraphRAG}): $x$-axis is ingestion cost (USD, log scale), $y$-axis is per-domain QA accuracy (\%), and bubble area encodes on-disk storage in GB. Per-(baseline,\,domain) numerical values are listed in Table~\ref{tab:efficiency_full}.}
    \label{fig:tradeoff}
\end{figure}

Per-domain ingestion cost varies by nearly two orders of magnitude across baselines (\$0.31--\$51.04), and on-disk storage spans 0.69--8.63 GB. Figure~\ref{fig:tradeoff} answers whether this expense translates into accuracy; per-(baseline,\,domain) values are tabulated in Appendix~\ref{appendix:efficiency_table}.

\textbf{Most baselines are Pareto-dominated by HippoRAG.}
\textsc{HippoRAG} sits on the cost--accuracy Pareto front in all four domains; \textsc{A-Mem}, \textsc{Mem0}, and \textsc{GraphRAG} are dominated in three or four of them. The \textsc{GraphRAG} case is informative: despite sharing similar ``graph + RAG'' structure as HippoRAG, it costs more and trails by 14--25 absolute accuracy points, indicating that the entity-centric graph construction in HippoRAG is the load-bearing design choice rather than graph augmentation in the abstract. The two ends of the front are \textsc{MemGPT} (cheap embedding-only ingest, accuracy capped at 22--35\%) and \textsc{Hindsight} (highest accuracy at 2--5$\times$ HippoRAG's cost), where the cost premium buys only 2--12 absolute accuracy points, concentrated on the smaller corpora.

\textbf{Storage does not track accuracy.}
The two largest on-disk footprints, \textsc{HippoRAG} and \textsc{A-Mem}, bracket the accuracy spectrum: HippoRAG sits on the Pareto front while A-Mem is dominated everywhere. \textsc{Hindsight} reaches the highest accuracy with a smaller footprint, and \textsc{MemGPT}'s compact 0.69 GB store yields mid-pack accuracy.

\textbf{The agent-memory paradigm is itself Pareto-dominated by raw retrieval.}
\textsc{BM25}'s ingestion is a one-time inverted-index build with no LLM or embedding calls (effectively \$0), yet it reaches 43.2\% cross-domain accuracy, strictly Pareto-dominating four of the five agent memory systems and surpassed only by \textsc{Hindsight}. Combined with Section~\ref{sec:failure_analysis}, this says the transformations memory systems perform during ingestion (\emph{e.g.}, fact extraction, graph construction, note rewriting) do not, on average, repay their cost in retrievability beyond what raw text already provides.

\subsection{Failure Analysis}
\label{sec:failure_analysis}

\begin{figure*}[t]
  \centering
  \includegraphics[width=\linewidth]{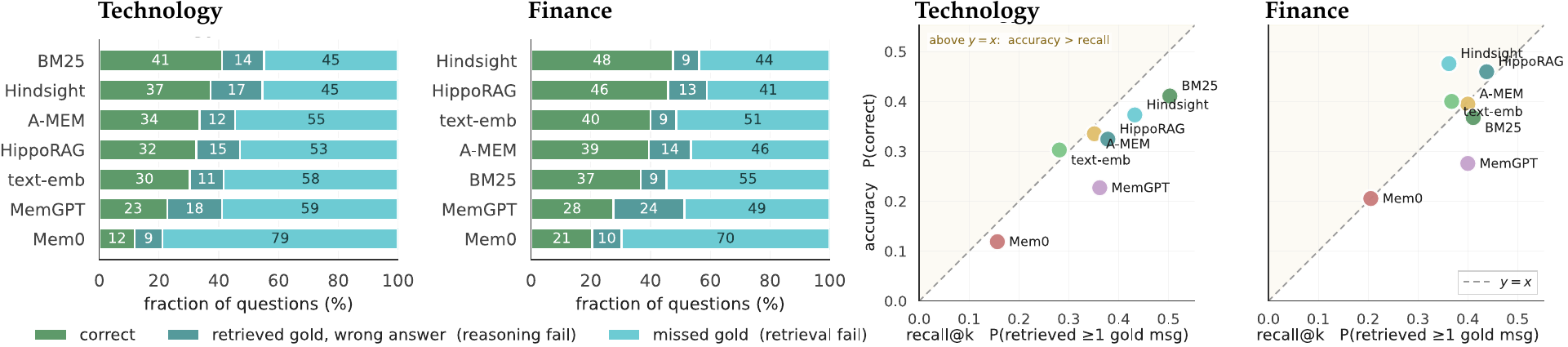}
  \vspace{-20pt}
  \caption{\textbf{(Left)} Failure-mode decomposition: each baseline's 185 non-abstention questions per domain split into \emph{correct}, \emph{reasoning} failure, and \emph{retrieval} failure.
    \textbf{(Right)} Retrieval recall vs.\ answer accuracy. Markers \emph{on} the diagonal are retrieval-bottlenecked; \emph{below} the diagonal indicates reasoning loss (gold surfaced but answered wrong); \emph{above} indicates the system answered correctly without the gold message, typically by aggregating other ingested context.
    }
  \label{fig:failure_analysis}
\end{figure*}

We dissect the mechanism behind the gaps above by attributing each error to either \emph{retrieval failure} (no gold supporting message) or \emph{reasoning failure} (gold retrieved but the LLM answered incorrectly).

\textbf{Retrieval, not reasoning, is the bottleneck (Q1, Q2).}
Across all baselines, retrieval failures account for 41--79\% of questions while reasoning failures stay narrowly bounded (7--24\%): given the gold message, current LLMs reliably synthesise a correct answer. Most baselines lie on $y{=}x$ in Figure~\ref{fig:failure_analysis}(Right), so accuracy scales near-linearly with recall. Two diagnostic deviations: \textsc{Hindsight} on Finance lies above the diagonal as its paraphrased notes aggregate context across messages, while \textsc{MemGPT} sits clearly below, retrieving adequately yet answering far worse, the signature of a representation that degrades downstream reasoning. This localizes the Q1/Q2 gaps to ingestion: structure and speaker context are lost before the LLM ever sees them.

\begin{wrapfigure}{r}{0.6\linewidth}
  \vspace{-1.2em}
  \centering
  \includegraphics[width=\linewidth]{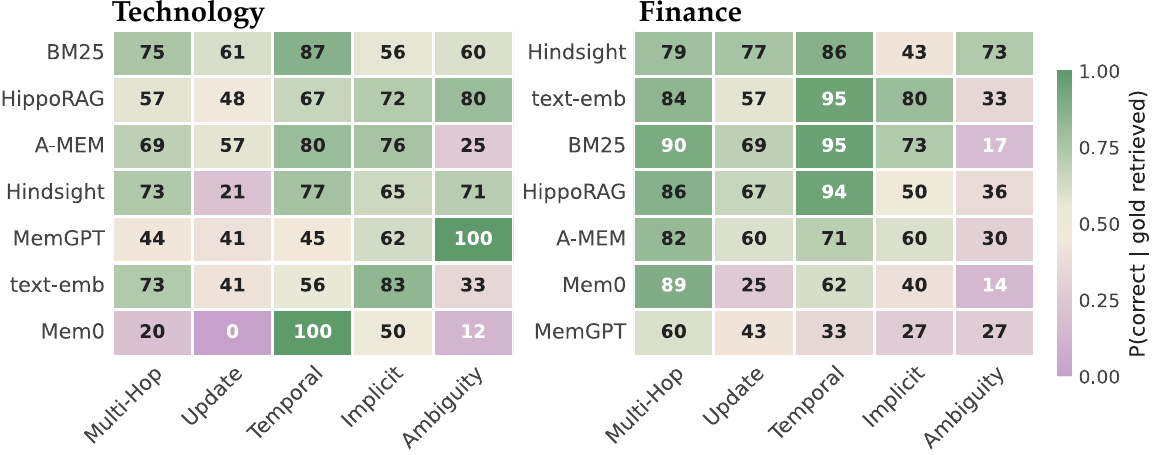}
  \vspace{-20pt}
  \caption{$P(\mathrm{correct}\mid\mathrm{gold\ retrieved})$ per
    (baseline, question type). Factoring out retriever quality
    isolates each memory representation's reasoning ability.}
  \label{fig:analysis_heatmap}
  \vspace{-0.8em}
\end{wrapfigure}

\textbf{Lexical shifts are the only failure that survives retrieval (Q3).}
Factoring out retriever quality (Figure~\ref{fig:analysis_heatmap}), \texttt{temporal} is uniformly high, effectively solved given retrieval, whereas \texttt{term\_ambiguity} stays below 40\% across nearly every cell, the only type where representation, not retrieval, is the wall. \textsc{BM25}, which feeds the LLM raw message text, attains the highest mean conditional accuracy in both domains, matching or exceeding every memory-augmented baseline. The transformations performed by memory systems therefore tend to \emph{reduce} rather than enhance the LLM's ability to reason from retrieved evidence.

\section{Conclusion}
We introduced \textbf{GroupMemBench}, a benchmark for agent memory in multi-party conversations. Our evaluation indicates that leading memory systems collapse across the benchmark and a bare \textsc{BM25} baseline matches or exceeds most of them, showing that current ingestion pipelines erase the important features (\emph{e.g.}, threaded conversation structure, lexical shifts) that group memory fundamentally depends on. Our finding highlights the need for more generalized agent memory systems for both dyadic and multi-party memory settings, and \textbf{GroupMemBench} drives valuable future improvements in multi-party memory solutions.




\bibliographystyle{unsrtnat}
\bibliography{custom}

\appendix

\section*{Appendix}
\section{Graph Schema}
\label{appendix:graph_schema}

\paragraph{Node types and attributes.}
The synthesis graph $\mathcal{G}$ contains four structural node types created
during graph construction and one dynamic node type added during message
generation:

\begin{itemize}[leftmargin=10pt, itemsep=1pt]
    \item \textbf{Domain} ($d$): a project. Stores the project's category
    tags (\emph{e.g.}, \emph{Quality Control}, \emph{Compliance}), which are used to
    induce \texttt{related\_to\_project} edges between sibling projects.
    \item \textbf{Topic} ($\tau$): a thematic decomposition of a domain.
    Topic identifiers are namespaced as \texttt{\$d::topic\_label\$} to avoid
    collisions across projects.
    \item \textbf{Phase} ($\phi$): a temporally bounded sub-problem of a topic.
    Each phase carries (i) a \emph{progress category} drawn from
    $\{\textsc{Risks},\,\textsc{Milestone},\,\textsc{Work Items}\}$, (ii) a
    \emph{status goal} drawn from a category-specific schema (\emph{e.g.},
    \textsc{Detected}/\textsc{Mitigated}/\textsc{Escalated} for Risks), (iii)
    a \emph{start} and \emph{target} date used for timestamp generation and
    progress tracking, (iv) a phase \emph{owner}, and (v) a synthetic
    documentation URL referenced by some prompts.
    \item \textbf{User} ($u$): a participant. Each user is assigned a
    persona $\mathcal{C}_u = (\textit{role},\textit{tone},\textit{style},
    \textit{expertise})$ where \emph{tone} and \emph{style} are drawn jointly
    \emph{without replacement} from the Cartesian product of the two
    vocabularies, so every user has a distinct voice. The full vocabularies
    are listed in Table~\ref{tab:persona_vocab}.
    \item \textbf{Message} ($m$): added at generation time. Stores the
    rendered text, the author, the timestamp, and the \emph{path type}
    (\texttt{post}, \texttt{reply}, or \texttt{cross\_project\_reply}) used
    to produce it.
\end{itemize}

\paragraph{Edges.}
The full set of static and dynamic relations is given in
Table~\ref{tab:relations}.  The two relations that most strongly shape the
synthetic discourse are \texttt{active\_in\_phase} (which restricts the
candidate authors of any message to the phase's assigned members) and
\texttt{related\_to\_project} (which is added between any two domains that
share at least one category tag and which is the only edge that the
cross-project handler walks across).

\paragraph{Persona vocabularies.}
Each user is assigned a persona along four axes: \emph{tone}, \emph{style},
\emph{expertise}, and \emph{role}. The vocabularies for the first three axes
are domain-agnostic and listed in Table~\ref{tab:persona_vocab}; the
\emph{role} vocabulary is domain-specific and drawn from the pool defined for
each project family in our configuration (\emph{e.g.}, for Manufacturing:
\emph{Production Manager}, \emph{Quality Engineer}, \emph{Maintenance
Engineer}, \emph{Supply Chain Manager}, \emph{Project Manager}). To ensure
that every user has a recognisable individual voice, $(\textit{tone},
\textit{style})$ pairs are sampled \emph{jointly without replacement} from
the Cartesian product of the tone and style vocabularies, while expertise
and role are sampled independently and uniformly. The tone, style, and
expertise values do more than label the user: each value is paired with a
concrete behavioural descriptor (\emph{e.g.}, the \emph{anecdotal} style instructs
the model to ground arguments in past experience without inventing project
codenames) that is injected verbatim into the system prompt during message
generation, as described in Appendix~\ref{appendix:prompt}.

\begin{table}[h]
\centering
\small
\caption{Persona vocabularies. Tone and style values are paired by sampling
without replacement so users have distinct $(\textit{tone},\textit{style})$
combinations.}
\label{tab:persona_vocab}
\begin{tabular}{l p{0.65\linewidth}}
\toprule
\textbf{Attribute} & \textbf{Values} \\
\midrule
Tone       & direct, casual, professional, empathetic, persuasive,
             humorous, skeptical, enthusiastic, anxious \\
Style      & concise, chatty, elaborative, boring, action-oriented,
             anecdotal \\
Expertise  & novice, intermediate, expert, specialist \\
Role       & domain-specific (\emph{e.g.}, for Manufacturing: Production Manager,
             Quality Engineer, Maintenance Engineer, Supply Chain Manager,
             Project Manager) \\
\bottomrule
\end{tabular}
\end{table}

\section{Path Sampling and Graph Walks}
\label{appendix:path_sampling}

\begin{table}[h]
\centering
\small
\caption{Edge relations in the synthesis graph $\mathcal{G}$.
``Static'' edges are created during graph construction; ``dynamic'' edges
are added incrementally as messages are generated.}
\label{tab:relations}
\begin{tabular}{l l l l}
\toprule
\textbf{Relation} & \textbf{Source} $\to$ \textbf{Target} & \textbf{Type} & \textbf{Purpose} \\
\midrule
\texttt{has\_topic}              & $d \to \tau$              & static  & domain owns topic \\
\texttt{has\_phase}              & $\tau \to \phi$           & static  & topic decomposes into phases \\
\texttt{belongs\_to\_project}    & $\phi \to \tau$           & static  & inverse pointer for traversal \\
\texttt{involved\_in}            & $u \to d$                 & static  & user is on the project \\
\texttt{has\_member}             & $d \to u$                 & static  & inverse member edge \\
\texttt{active\_in\_phase}       & $u \to \phi$              & static  & user is active in the phase \\
\texttt{has\_phase\_member}      & $\phi \to u$              & static  & inverse phase-membership \\
\texttt{related\_to\_project}    & $d \to d'$                & static  & shared category bridge (cross-project) \\
\midrule
\texttt{authored\_by}            & $m \to u$                 & dynamic & message authorship \\
\texttt{posted\_in}              & $m \to \phi$              & dynamic & message anchored in phase \\
\texttt{reply\_to}               & $m \to m'$                & dynamic & threaded reply edge \\
\bottomrule
\end{tabular}
\end{table}

At each generation step we first sample a path type from the mixture
$\bigl(p_{\text{post}},\,p_{\text{reply}},\,p_{\text{cross}}\bigr)
= (0.20,\,0.60,\,0.20)$, with two deterministic overrides: (i) phases that
have not yet received any message are forced into the \texttt{post} branch,
and (ii) a phase whose recent posts are all open questions without engagement
(streak $\geq 3$, see ``question cooldown'' in
Appendix~\ref{appendix:hyperparams}) is forced into the \texttt{reply}
branch.  The chosen handler then walks $\mathcal{G}$ as follows.

\paragraph{Channel post ($d \to \tau \to \phi \to u$).}
Domains with no messages so far are sampled with probability $0.8$ to ensure
breadth coverage; otherwise a domain is chosen uniformly. From the domain we
walk a uniform \texttt{has\_topic} edge, then a \texttt{has\_phase} edge
weighted by the \emph{post-need} score
\[
w_{\text{post}}(\phi) =
\begin{cases}
10.0 & \text{if } \phi \text{ has no messages,} \\
8.0  & \text{if } \phi \text{ has no posts,} \\
\max\bigl(1.0,\;5.0 - 1.5\,n_{\text{posts}}(\phi)\bigr) +
2\sin\bigl(\pi\,p(\phi)\bigr) & \text{otherwise,}
\end{cases}
\]
where $p(\phi) \in [0,1]$ is the phase's temporal progress.  The author $u$
is then sampled uniformly from the phase's \texttt{active\_in\_phase}
members.  The post type is conditioned on the existing message count:
$\{$kickoff, milestone$\}$ for empty phases, $\{$question, update,
coordination$\}$ at low volume, and the full set
$\{$update, question, blocker, decision, milestone, escalation,
coordination$\}$ thereafter.

\paragraph{Threaded reply ($d \to \tau \to \phi \to m \to u$).}
A domain is sampled inversely proportional to its current message count so
that high-volume projects do not monopolise replies.  Within the domain we
walk to a phase under the \emph{phase-heat} score
\[
w_{\text{heat}}(\phi) = 1 + 0.5\,\min(n,10) + 3.0\,n_{\text{unanswered}}
                       + 2\sin\bigl(\pi\,p(\phi)\bigr),
\]
where $n_{\text{unanswered}}$ counts posts that have received no reply.  The
parent message $m$ is then chosen from the phase's history with weight
\[
w(m) \;=\; \underbrace{(1+r_m)^{1.3} \cdot \mathbb{1}[r_m{<}4 \,\lor\, 0.35]}_{\text{preferential attachment + saturation}}
        \;\cdot\; \underbrace{\max\!\left(0.15,\;1-\tfrac{a_m}{20}\right)}_{\text{recency decay}}
        \;\cdot\; \underbrace{\bigl(2.0 \text{ if } m \text{ is a post else } 1.0\bigr)}_{\text{post-over-reply bias}},
\]
where $r_m$ is the current reply count of $m$ and $a_m$ its age in messages.
The responder is drawn uniformly from active phase members excluding the
parent author and the last three speakers when possible.

\paragraph{Cross-project bridge ($u \!\to\! \phi \!\to\! \tau \!\to\! d \!\xrightarrow{\text{rel.}}\! d' \!\to\! \tau' \!\to\! \phi' \!\to\! u'$).}
We anchor at a uniformly sampled user $u$, follow their
\texttt{active\_in\_phase} edge, climb back through topic and domain, cross
exactly one \texttt{related\_to\_project} edge to a sibling project $d'$, and
descend to a target phase $\phi'$ in $d'$ chosen by the
\emph{active-thread} score
$w_{\text{thread}}(\phi) = 1 + 0.5\min(n_\phi,15)$.  The target user $u'$ is
preferentially someone with the same role as $u$ in $\phi'$, and the walk
is rejected if $\phi'$ starts strictly before $\phi$ (to keep the temporal
direction of cross-project influence consistent).

\paragraph{Situational context assembly.}
After the path is fixed we assemble the situational context $\mathcal{S}_u$
by concatenating: (i) the most recent $W=50$ messages of the target phase,
optionally compressed by sampling 3 short snippets from any older window;
(ii) the phase memory artefact described in
Appendix~\ref{appendix:hyperparams} (sub-issue + ledger); and, with
probability $0.25$, (iii) a one-sentence cross-reference snippet drawn
uniformly from the recent history of a sibling phase in the same project.
$\mathcal{S}_u$ is then deduplicated against itself with a sequence-matcher
threshold of $0.50$ to suppress paraphrastic echoes.

\section{Prompt Templates}
\label{appendix:prompt}

All prompts share a common \emph{system} prompt that fixes the speaker's
voice, and one of four \emph{user} prompt templates that fixes the
conversational scenario.  Templates are abbreviated below; full strings,
including the persona-conditioned behavioural descriptors injected into the
\textsc{Voice} block, are released with the code.

\paragraph{System prompt (per-user, persona-conditioned).}
\begin{quote}\small\ttfamily
You are a \{role\} on a Microsoft Teams channel.\\
\textsc{Voice}: tone=\{tone\_desc\}; style=\{style\_desc\};
expertise=\{expertise\_desc\}.\\
\textsc{Length}: replies 1--3 sentences ($\le$15 words each); posts 2--4
sentences. Never write paragraphs unless asked.\\
\textsc{Format}: \{style-specific formatting rule\}.\\
\textsc{Rules}: stay in character; vary openings; never repeat a previous
opening or anecdote; do not parrot the last few messages.
\end{quote}

\paragraph{Channel-post template ($T_{\text{post}}$).}
\begin{quote}\small\ttfamily
\textsc{Background}: \{user, persona, enriched phase background\}.\\
\textsc{Phase}: \{phase\_name\} ($\{stage\}$ stage), status goal=\{status\},
target=\{date\}; stage guidance=\{...\}.\\
\textsc{Task}: \{post\_type\_purpose\} (kickoff / update / question /
blocker / decision / milestone / escalation / coordination), in the context
of \{post\_focus\} for the \{phase\_name\} phase.\\
\textsc{Voice}: tone, style, expertise descriptors.\\
\textsc{Constraints}: 2--4 sentences; pick one concrete thing; do not
announce phase start or percentage complete; phrasing rules
(banned phrases / openings, length target, optional messiness instruction);
optional doc-URL instruction.
\end{quote}

\paragraph{Threaded-reply template ($T_{\text{reply}}$).}
\begin{quote}\small\ttfamily
\textsc{Background}, \textsc{Phase}: as above.\\
\textsc{Channel history}: \{situational context $\mathcal{S}_u$\}.\\
\textsc{Task}: write a reply to \{parent\_message\} in the \{project\}
channel; build on what has been said, address the original question, push
toward closure if late in the phase.\\
\textsc{Voice}, \textsc{Mention}: optional \texttt{@}\{parent\_author\}.\\
\textsc{Constraints}: phrasing rules; with probability $0.30$ a hard
``ULTRA-SHORT 1--6 words'' override; with probability $0.25$ a mandatory
cross-reference to a sibling phase; closure instruction emitted when phase
progress $> 0.8$.
\end{quote}

\paragraph{Cross-project role-to-role template ($T_{\text{cross}}$).}
\begin{quote}\small\ttfamily
\textsc{Background}: \{user, persona, source phase\}.\\
\textsc{Recipient role}: same as author's role.\\
\textsc{Original post}: \{first\_post\_in\_target\_phase\}.\\
\textsc{Thread context}: \{$\mathcal{S}_u$ from target phase\}.\\
\textsc{Task}: bring an experience-based perspective from the source
project; share a transferable insight, ask a thoughtful question, or
suggest an approach that helps move the target phase toward its
status goal; reference earlier messages naturally.\\
\textsc{Voice}, phrasing rules, optional doc-URL, closure instruction.
\end{quote}

\paragraph{Disagreement template ($T_{\text{disagree}}$, used in $25\%$ of replies).}
\begin{quote}\small\ttfamily
\textsc{Task}: write a reply that pushes back on a point in the
conversation; choose ONE of:
challenge feasibility, question assumptions, propose alternative,
flag overlooked risk, push back on scope, cite counter-evidence.
The disagreement must be substantive (not ``I agree but...'').
\end{quote}

\paragraph{Noise template ($T_{\text{noise}}$, used with prob.\ $0.05$).}
\begin{quote}\small\ttfamily
\textsc{Task}: write a message that contains exactly one realistic
mistake, drawn uniformly from
\{wrong detail, misunderstanding, stale info, off-topic drift,
premature assumption, wrong audience\}.  The mistake must be
\emph{subtle}; do not signal uncertainty; write with full confidence.
\end{quote}

\paragraph{Phase-memory prompts.}
Two auxiliary prompts are issued to the same model independently of the
message-generation prompts:
(i) a \emph{sub-issue prompt} called once per phase asking for
$K{=}6$ distinct, non-overlapping sub-issues a real cross-functional team
would surface, and
(ii) a \emph{ledger prompt} called every $L{=}8$ new messages that rewrites
a $\le 180$-word summary into three sections,
\textsc{Decided} / \textsc{Open Questions} (with repeat counts) /
\textsc{Pending Commitments}, and is required to migrate any newly answered
question from \textsc{Open} to \textsc{Decided}.

\section{Hyperparameters and Realism Knobs}
\label{appendix:hyperparams}

Table~\ref{tab:hparams} lists the values used to produce the corpus
described in this paper.  All values are taken from the released code and
are exposed as a single \texttt{POLICY} dict so that they can be
re-tuned in one place.

\begin{table}[h]
\centering
\small
\caption{Generation hyperparameters.}
\label{tab:hparams}
\begin{tabular}{l l l}
\toprule
\textbf{Group} & \textbf{Parameter} & \textbf{Value} \\
\midrule
Path sampling
  & $p_{\text{post}}$ : $p_{\text{reply}}$ : $p_{\text{cross}}$ & $0.20:0.60:0.20$ \\
  & Empty-phase post override probability                       & $0.80$ \\
  & Force-reply on question streak                              & $\geq 3$ \\
  & Reply target post bias                                      & $\times 2.0$ \\
  & Reply target preferential-attachment exponent               & $1.3$ \\
  & Reply target recency window / floor                         & $20$ msgs $/$ $0.15$ \\
\midrule
Context window
  & Recent verbatim window $W$                                  & $50$ messages \\
  & Older-context summary snippets                              & $3$ \\
  & Deduplication threshold                                     & $0.50$ \\
\midrule
Realism perturbations
  & Noise rate                                                  & $0.05$ \\
  & Disagreement rate (replies only)                            & $0.25$ \\
  & Stylistic-imperfection rate                                 & $0.20$ \\
  & Ultra-short reply rate                                      & $0.30$ \\
  & Tone jitter rate                                            & $0.10$--$0.20$ \\
  & URL inclusion rate                                          & $0.10$ (post: $0.35$) \\
  & Cross-reference rate                                        & $0.25$ \\
  & Forced decision-change eligibility                          & $0.3 \le p(\phi) \le 0.8$ and $\geq 4$ msgs \\
\midrule
Phase memory
  & Sub-issues per phase                                        & $6$ \\
  & Sub-issue steering rate                                     & $0.70$ \\
  & Ledger refresh interval                                     & $8$ new messages \\
  & Ledger context window                                       & $16$ messages \\
\midrule
Timestamps
  & Phase duration                                              & $10$ days \\
  & Inter-arrival distribution                                  & exponential
                                                                  (Poisson process) \\
  & Late-stage message cap                                      & $10$ msgs after $p(\phi){\geq}0.95$ \\
\midrule
Corpus scale
  & Target messages (full run)                                  & $30{,}000$ \\
  & Topics per project                                          & $3$ \\
  & Phases per topic                                            & $5$ \\
  & Users per project                                           & $4$ (out of $6$--$10$) \\
  & Backbone language model                                     & GPT-5 \\
\bottomrule
\end{tabular}
\end{table}
\section{Evaluation Rubrics for Synthetic Conversation}
\label{appendix:geval}
For all synthetic group chat data, we report quality using six core metrics, each scored by LLM-as-Judge protocols (\emph{e.g.}, G-Eval) on a scale from 1 to 5, with higher values indicating better performance:
\begin{itemize}[leftmargin=1em,itemsep=0.5pt]
\item \textbf{Naturalness:} Measures how authentic and human-like the conversation appears. High scores indicate messages resemble real workplace communication, with natural phrasing and plausible imperfections.
\item \textbf{Coherence:} Assesses the logical flow and connectivity between messages. Coherent conversations maintain topic continuity and clear transitions, avoiding abrupt or confusing shifts.
\item \textbf{Diversity:} Captures the variety in communication styles, topics, and participant contributions. High diversity reflects a mix of message lengths, tones, and perspectives, avoiding repetitive exchanges.
\item \textbf{Contextual Relevance:} Evaluates how well each message relates to the shared context and ongoing discussion. Relevant conversations consistently reference prior messages, project details, and team objectives.
\item \textbf{Momentum:} Measures the degree to which the conversation progresses toward goals or resolutions. High momentum indicates that messages drive the discussion forward, address blockers, and facilitate decision-making.
\item \textbf{Engagingness:} Assesses how interactive and stimulating the conversation is for participants. Engaging conversations feature active participation, questions, acknowledgments, and responses that encourage further dialogue.
\item \textbf{Overall Avg.:} The arithmetic mean of the above metrics, providing a single summary score for overall conversation quality.
\end{itemize}
These metrics collectively capture the realism, effectiveness, and collaborative dynamics of synthetic workplace conversations, enabling nuanced evaluation and comparison across datasets.

\section{Dataset Statistics}
\label{appendix:dataset_stats}

Table~\ref{tab:conversation} reports per-domain corpus statistics for the
synthesized \textsc{GroupMemBench} conversations. Each domain is realized as
a collection of \emph{projects} (one Teams-style channel per project),
which are in turn decomposed into topics and time-bounded phases following
the graph schema of Section~\ref{sec:dataset_construction}. Within every
phase, individual messages are tagged with a \emph{path type}
(\emph{post}, \emph{reply}, or \emph{cross-project reply}), and may
additionally be flagged as a \emph{noise} or \emph{decision} message.

Three properties of the corpus are intentional. First, the four domains are
\textbf{matched on message count} ($30{,}000$ messages per domain), so any
cross-domain difference in downstream accuracy or cost cannot be attributed
to corpus-size imbalance. Second, the per-path-type proportions are stable
across domains and closely track the path-sampling mixture
$(p_{\text{post}}, p_{\text{reply}}, p_{\text{cross}}) = (0.20, 0.60, 0.20)$:
posts and replies remain near a $1{:}3$ ratio, with an additional
$\sim$10\% of messages crossing project boundaries. Third, the participant
and role counts vary by design, ranging from a leaner Healthcare setting
($6$ participants, $4$ roles) to a richer Technology setting ($18$
participants, $6$ roles), so memory-evaluation tasks see a spectrum of
organizational complexity rather than a single homogeneous setting.

\begin{table}[h]
\caption{Per-domain statistics of the synthesized group conversations.}
\label{tab:conversation}
\centering
\small
\setlength{\tabcolsep}{3pt}
\begin{tabular}{@{}lcccc@{}}
\toprule
\textbf{Attribute} & \textbf{Technology} & \textbf{Finance} & \textbf{Healthcare} & \textbf{Manufacturing} \\
\midrule
\# Projects             & 7      & 6      & 10     & 10 \\
\# Topics               & 35     & 30     & 30     & 30 \\
\# Phases               & 74     & 66     & 79     & 76 \\
\# Participants         & 18     & 12     & 6      & 9  \\
\# Roles                & 6      & 7      & 4      & 5  \\
\# Posts                & 6{,}599 & 6{,}764 & 7{,}043 & 6{,}816 \\
\# Replies              & 19{,}897 & 19{,}722 & 20{,}099 & 20{,}048 \\
\# Cross-project replies & 3{,}504 & 3{,}514 & 2{,}858 & 3{,}136 \\
\# Noise messages       & 1{,}503 & 1{,}535 & 1{,}510 & 1{,}542 \\
\# Decision points      & 20{,}788 & 19{,}920 & 10{,}301 & 7{,}681 \\
\# Decision changes     & 74     & 66     & 113    & 99 \\
Total messages          & 30{,}000 & 30{,}000 & 30{,}000 & 30{,}000 \\
\bottomrule
\end{tabular}
\end{table}

\section{Per-Domain Results}
\label{appendix:per_domain}

Tables~\ref{tab:per_domain_technology}--\ref{tab:per_domain_manufacturing} report the per-domain accuracy of each baseline on the four domains of GroupMemBench (Technology, Finance, Healthcare, Manufacturing). The cross-domain micro-averages summarized in Table~\ref{tab:main_results} are computed over the union of the filtered evaluation set across these four domains.


\begin{table}[t]
\caption{Per-domain accuracy (\%) on the \textbf{Technology} subset of GroupMemBench. Best per column in \textbf{bold}, second-best \underline{underlined}.}
\label{tab:per_domain_technology}
\centering
\footnotesize
\setlength{\tabcolsep}{4pt}
\renewcommand{\arraystretch}{1.25}
\resizebox{\linewidth}{!}{%
\begin{tabular}{l ccccccc}
\toprule
\textbf{Method} & \textbf{Multi-Hop} & \textbf{Update} & \textbf{Ambiguity} & \textbf{Implicit} & \textbf{Temporal} & \textbf{Abstention} & \textbf{Average} \\
\midrule
\rowcolor{raghdr} \multicolumn{8}{l}{\textit{RAG-based Methods}} \\
BM25                                 & \textbf{43.90} & \underline{30.60} & 23.30          & 35.70          & \textbf{73.00} & 78.60          & \textbf{46.00} \\
text-embedding-3-large          & 31.70          & 25.00          & 20.90          & 46.40          & 32.40          & \underline{82.10} & 37.10 \\
GraphRAG~\citep{edge2024local}                          & 19.50          & 19.40          & 16.30          & 17.90          & \phantom{0}5.40 & 78.60          & 23.90 \\
\midrule
\rowcolor{agenthdr} \multicolumn{8}{l}{\textit{Agent Memory Systems}} \\
Mem0~\citep{chhikara2025mem0}                           & \phantom{0}7.30 & \phantom{0}0.00 & 11.60         & 17.90          & 24.30          & \textbf{89.30} & 22.10 \\
MemGPT~\citep{packer2023memgpt}                         & 24.39          & 25.00          & 16.28          & 35.71          & 16.22          & 75.00          & 29.58 \\
A-Mem~\citep{xu2025mem}                                 & \underline{36.60} & \textbf{33.30} & 25.60          & \textbf{53.60} & 24.30          & 67.90          & 38.00 \\
HippoRAG~\citep{gutierrez2024hipporag}                  & 26.80          & \textbf{33.30} & \underline{27.90} & \underline{50.00} & 29.70          & 78.60          & 38.50 \\
Hindsight~\citep{latimer2025hindsight}                  & \underline{36.60} & 11.10          & \textbf{37.20} & \underline{50.00} & \underline{54.10} & 71.40          & \underline{41.80} \\
\bottomrule
\end{tabular}%
}
\end{table}

\begin{table}[t]
\caption{Per-domain accuracy (\%) on the \textbf{Finance} subset of GroupMemBench. Best per column in \textbf{bold}, second-best \underline{underlined}.}
\label{tab:per_domain_finance}
\centering
\footnotesize
\setlength{\tabcolsep}{4pt}
\renewcommand{\arraystretch}{1.25}
\resizebox{\linewidth}{!}{%
\begin{tabular}{l ccccccc}
\toprule
\textbf{Method} & \textbf{Multi-Hop} & \textbf{Update} & \textbf{Ambiguity} & \textbf{Implicit} & \textbf{Temporal} & \textbf{Abstention} & \textbf{Average} \\
\midrule
\rowcolor{raghdr} \multicolumn{8}{l}{\textit{RAG-based Methods}} \\
BM25                                & 47.90          & \underline{34.40} & 11.10          & \underline{53.30} & 46.70          & 75.90          & 42.10 \\
text-embedding-3-large          & 52.10          & 31.20          & 22.20          & \textbf{60.00} & 44.40          & 72.40          & 44.40 \\
GraphRAG~\citep{edge2024local}                          & 22.90          & 12.50          & 26.70          & \phantom{0}6.70 & \phantom{0}6.70 & 72.40          & 24.30 \\
\midrule
\rowcolor{agenthdr} \multicolumn{8}{l}{\textit{Agent Memory Systems}} \\
Mem0~\citep{chhikara2025mem0}                           & 31.20          & \phantom{0}6.20 & \phantom{0}8.90 & 20.00         & 31.10          & \textbf{86.20} & 29.40 \\
MemGPT~\citep{packer2023memgpt}                         & 41.67          & 18.75          & 28.89          & 26.67          & 17.78          & \underline{79.31} & 34.58 \\
A-Mem~\citep{xu2025mem}                                 & \underline{56.20} & 31.20          & 33.30          & 46.70          & 31.10          & 69.00          & 43.50 \\
HippoRAG~\citep{gutierrez2024hipporag}                  & \textbf{58.30} & \textbf{37.50} & \underline{40.00} & 33.30          & \underline{48.90} & 69.00          & \underline{49.10} \\
Hindsight~\citep{latimer2025hindsight}                  & 50.00          & 31.20          & \textbf{44.40} & 33.30          & \textbf{64.40} & 79.30          & \textbf{51.90} \\
\bottomrule
\end{tabular}%
}
\end{table}

\begin{table}[t]
\caption{Per-domain accuracy (\%) on the \textbf{Healthcare} subset of GroupMemBench. Best per column in \textbf{bold}, second-best \underline{underlined}. The Ambiguity and Implicit columns include 0\,\% / 50\,\% ties across many baselines, reflecting the small per-cell sample size of those question types in this domain.}
\label{tab:per_domain_healthcare}
\centering
\footnotesize
\setlength{\tabcolsep}{4pt}
\renewcommand{\arraystretch}{1.25}
\resizebox{\linewidth}{!}{%
\begin{tabular}{l ccccccc}
\toprule
\textbf{Method} & \textbf{Multi-Hop} & \textbf{Update} & \textbf{Ambiguity} & \textbf{Implicit} & \textbf{Temporal} & \textbf{Abstention} & \textbf{Average} \\
\midrule
\rowcolor{raghdr} \multicolumn{8}{l}{\textit{RAG-based Methods}} \\
BM25                                  & \underline{37.50} & \underline{11.80} & \phantom{0}0.00 & \textbf{100.00} & \textbf{45.90} & \textbf{89.50} & \underline{43.10} \\
text-embedding-3-large          & 27.10          & \textbf{17.60} & \textbf{14.30} & \underline{50.00} & 10.80          & 68.40          & 26.90 \\
GraphRAG~\citep{edge2024local}                          & \phantom{0}2.10 & \phantom{0}5.90 & \phantom{0}0.00 & \underline{50.00} & \phantom{0}2.70 & 57.90          & 11.50 \\
\midrule
\rowcolor{agenthdr} \multicolumn{8}{l}{\textit{Agent Memory Systems}} \\
Mem0~\citep{chhikara2025mem0}                           & 12.50          & \phantom{0}5.90 & \phantom{0}0.00 & \underline{50.00} & \phantom{0}0.00 & 73.70         & 16.90 \\
MemGPT~\citep{packer2023memgpt}                         & 12.50          & \textbf{17.60} & \phantom{0}0.00 & \phantom{0}0.00 & \phantom{0}5.60 & \textbf{89.50} & 21.70 \\
A-Mem~\citep{xu2025mem}                                 & 22.90          & \underline{11.80} & \phantom{0}0.00 & \underline{50.00} & 10.80          & 68.40          & 23.80 \\
HippoRAG~\citep{gutierrez2024hipporag}                  & 35.40          & \textbf{17.60} & \phantom{0}0.00 & \textbf{100.00} & \underline{27.00} & 73.70          & 35.40 \\
Hindsight~\citep{latimer2025hindsight}                  & \textbf{45.80} & \textbf{17.60} & \textbf{14.30} & \underline{50.00} & \textbf{45.90} & \underline{78.90} & \textbf{45.40} \\
\bottomrule
\end{tabular}%
}
\end{table}

\begin{table}[t]
\caption{Per-domain accuracy (\%) on the \textbf{Manufacturing} subset of GroupMemBench. Best per column in \textbf{bold}, second-best \underline{underlined}.}
\label{tab:per_domain_manufacturing}
\centering
\footnotesize
\setlength{\tabcolsep}{4pt}
\renewcommand{\arraystretch}{1.25}
\resizebox{\linewidth}{!}{%
\begin{tabular}{l ccccccc}
\toprule
\textbf{Method} & \textbf{Multi-Hop} & \textbf{Update} & \textbf{Ambiguity} & \textbf{Implicit} & \textbf{Temporal} & \textbf{Abstention} & \textbf{Average} \\
\midrule
\rowcolor{raghdr} \multicolumn{8}{l}{\textit{RAG-based Methods}} \\
BM25                                  & 31.10          & \textbf{13.60} & \phantom{0}0.00 & \phantom{0}0.00 & \textbf{55.80} & 72.70          & \underline{41.10} \\
text-embedding-3-large          & \underline{33.30} & \textbf{13.60} & \textbf{27.30} & \phantom{0}0.00 & 39.50          & \underline{75.80} & 39.90 \\
GraphRAG~\citep{edge2024local}                          & \phantom{0}4.40 & \textbf{13.60} & \underline{18.20} & \phantom{0}0.00 & \phantom{0}7.00 & 57.60          & 18.40 \\
\midrule
\rowcolor{agenthdr} \multicolumn{8}{l}{\textit{Agent Memory Systems}} \\
Mem0~\citep{chhikara2025mem0}                           & \textbf{35.60} & \underline{\phantom{0}9.10} & \textbf{27.30} & \textbf{25.00} & \phantom{0}9.30 & \textbf{78.80} & 32.90 \\
MemGPT~\citep{packer2023memgpt}                         & 11.10          & \phantom{0}4.50 & \underline{18.20} & \phantom{0}0.00 & \phantom{0}9.30 & 72.70          & 22.80 \\
A-Mem~\citep{xu2025mem}                                 & 24.40          & \phantom{0}0.00 & \underline{18.20} & \phantom{0}0.00 & 25.60          & 66.70          & 29.10 \\
HippoRAG~\citep{gutierrez2024hipporag}                  & \textbf{35.60} & \underline{\phantom{0}9.10} & \underline{18.20} & \phantom{0}0.00 & 11.60          & \textbf{78.80} & 32.30 \\
Hindsight~\citep{latimer2025hindsight}                  & \textbf{35.60} & \underline{\phantom{0}9.10} & \textbf{27.30} & \phantom{0}0.00 & \underline{53.50} & \textbf{78.80} & \textbf{44.30} \\
\bottomrule
\end{tabular}%
}
\end{table}

\section{Per-Domain Ingestion Cost and Storage}
\label{appendix:efficiency_table}

Table~\ref{tab:efficiency_full} reports the per-domain ingestion cost (USD) and on-disk storage (GB) summarized in the bubble plot of Figure~\ref{fig:tradeoff}. Costs are computed from per-call token counts collected by a LiteLLM proxy and converted to USD using OpenAI list prices (\$0.15 / \$0.60 per 1M input/output tokens for \texttt{gpt-4o-mini}; \$0.13 per 1M tokens for \texttt{text-embedding-3-large}), summed across all successful LLM and embedding calls during ingestion. Query-time calls are excluded throughout. \textsc{MemGPT}'s archival ingest issues only embedding calls, so its LLM input/output token contributions are zero by design. Storage is measured directly with \texttt{du -sb} on each baseline's per-domain store directory; \textsc{Hindsight} and \textsc{MemGPT} share a single Postgres volume across domains (approximately 10.6 GB and 2.77 GB respectively), which we apportion to each domain in proportion to a per-domain bookkeeping counter the system itself reports---node + link counts for \textsc{Hindsight}, archival passage counts for \textsc{MemGPT}.


\begin{table}[t]
\caption{Per-domain ingestion cost (USD) and on-disk storage (GB) for the six baselines used in the trade-off analysis, corresponding to the bubble plot in Figure~\ref{fig:tradeoff}. Costs aggregate all LLM and embedding calls during ingestion (query-time calls excluded). The \textbf{Avg.}\ column is the unweighted mean across the four domains.}
\label{tab:efficiency_full}
\centering
\footnotesize
\setlength{\tabcolsep}{4pt}
\renewcommand{\arraystretch}{1.20}
\resizebox{\linewidth}{!}{%
\begin{tabular}{l rrrrr rrrrr}
\toprule
& \multicolumn{5}{c}{\textbf{Ingestion Cost (USD)}} & \multicolumn{5}{c}{\textbf{Storage (GB)}} \\
\cmidrule(lr){2-6} \cmidrule(lr){7-11}
\textbf{Method} & \textbf{Tech.} & \textbf{Fin.} & \textbf{Healthc.} & \textbf{Manuf.} & \textbf{Avg.}
                & \textbf{Tech.} & \textbf{Fin.} & \textbf{Healthc.} & \textbf{Manuf.} & \textbf{Avg.} \\
\midrule
Mem0~\citep{chhikara2025mem0}             & 16.82 & 34.92 & 10.99 & 10.89 & 18.41 & 0.75 & 1.21 & 1.00 & 1.01 & 0.99 \\
MemGPT~\citep{packer2023memgpt}           &  0.54 &  0.56 &  0.32 &  0.31 &  0.43 & 0.69 & 0.69 & 0.69 & 0.69 & 0.69 \\
HippoRAG~\citep{gutierrez2024hipporag}    &  9.32 & 10.01 &  7.66 &  7.55 &  8.64 & 6.94 & 6.98 & 4.45 & 4.68 & 5.76 \\
A-Mem~\citep{xu2025mem}                   & 39.44 & 42.66 & 11.14 & 11.28 & 26.13 & 7.53 & 8.63 & 5.12 & 5.13 & 6.60 \\
Hindsight~\citep{latimer2025hindsight}    & 21.81 & 51.04 & 27.95 & 27.60 & 32.10 & 5.26 & 5.35 & 3.34 & 3.04 & 4.25 \\
GraphRAG~\citep{edge2024local}            & 15.26 & 15.82 &  9.63 &  9.32 & 12.51 & 3.29 & 3.24 & 2.56 & 2.90 & 3.00 \\
\bottomrule
\end{tabular}%
}
\end{table}

\section{LLM Judge Configuration and Reliability}
\label{appendix:llm_judge}

All accuracy numbers in the main text are produced by a single LLM judge that compares each agent's answer to a gold reference and emits a binary verdict. This appendix specifies the judge's configuration, the prompt and output parsing rule, and a manual reliability check.

\paragraph{Judge configuration.}
The judge is GPT-5 served via the Azure OpenAI API (deployment \texttt{gpt-5}, API version \texttt{2024-02-15-preview}), with \texttt{max\_tokens}\,$=$\,2048 and \texttt{temperature}\,$=$\,1.0. Each (question, gold answer, predicted answer) triple is judged once; we do not use self-consistency or multi-sample voting.

\paragraph{Judge system prompt.}
The system message instructs the model to act as a strict equivalence judge, allow paraphrastic matches, and emit a structured verdict:
\begin{quote}\small\ttfamily
You are a strict judge evaluating whether an agent's answer matches the gold answer for a question.\\
Consider paraphrases correct if they have the same meaning as the gold answer.\\
First provide a brief reasoning paragraph. Then provide the final judgment on a new line using the format:\\
Final: Correct\\
or\\
Final: Incorrect
\end{quote}
The user message of each call contains the question, gold answer, and the agent's predicted answer concatenated under fixed headers.

\paragraph{Output parsing.}
The judge's reasoning is free-form, so we extract the verdict deterministically from the structured suffix. The parser scans the response for the last line beginning with \texttt{Final:} or \texttt{Final answer:}, lower-cases the text after the prefix, and applies the following precedence:
\begin{enumerate}[leftmargin=18pt, itemsep=1pt]
    \item If the suffix contains any of \texttt{incorrect}, \texttt{wrong}, or \texttt{not correct}, the verdict is \textbf{Incorrect}.
    \item Otherwise, if the suffix contains \texttt{correct}, the verdict is \textbf{Correct}.
    \item Otherwise, the verdict is recorded as \textbf{Unclear} and excluded from the accuracy denominator.
\end{enumerate}
Negative phrases are checked first because \texttt{not correct} is a substring of the positive trigger; the implementation is in \texttt{eval\_lib.py} (lines 146--152).

\paragraph{Reliability check.}
We manually re-examined 100 (question, gold answer, predicted answer, judge verdict) tuples sampled from the main evaluation run, covering all four domains and all six question types. The LLM judge agreed with our verdict on 99 of 100 items (99\%). The single disagreement was a false negative in which the agent's answer correctly contained the gold content but added extra material beyond what the gold specified; the judge interpreted the additional content as inconsistent with the gold answer and returned \emph{Incorrect}, where we would have accepted the response. We did not encounter any false positives in the sample, indicating that the judge errs slightly on the strict side rather than being lenient.

\paragraph{Implications for the main results.}
A 99\% agreement rate on this sample, combined with the strict-side bias of the lone disagreement, means that the reported accuracies are if anything mild lower bounds on the true correctness rate. Because the bias is uniform across baselines (no system is favoured by the judge's strictness), the cross-baseline ranking in Table~\ref{tab:main_results} is unaffected.

\section{Worked example: information loss during memory ingestion}
\label{app:information-loss}

We trace one Finance/user-implicit question through all eight
baselines---from BM25 (no semantic processing during ingestion) to
graphrag (per-domain knowledge graph + community summaries)---to
show concretely how an ingestion-time rewrite can either preserve or
erase the very fields the question depends on. The original chat
message that contains the answer is reproduced once; we then show
each baseline's top retrievals and the gpt-5 QA agent's answer,
grouped by whether that answer matches the gold.

\begin{tcolorbox}[
  title={\textbf{Question (Finance, user-implicit)}},
  colback=blue!4, colframe=blue!55!black,
  fonttitle=\bfseries,
  breakable,
]
\textbf{Asking user:} \texttt{User\_13} (Compliance Officer)\\
\textbf{Question:} ``Who do I need aligned on formatting rules for the
mitigation plan in the \emph{Risk: Formatting Inconsistencies} phase?''\\
\textbf{Gold answer:} \emph{Finance and Data Engineering.}
\end{tcolorbox}

\begin{tcolorbox}[
  title={\textbf{Ground truth in the conversation}\newline(\texttt{Msg\_28294}, written by the asker himself)},
  colback=gray!6, colframe=gray!55!black,
  fonttitle=\bfseries,
  breakable,
]
\textbf{Author:} \texttt{User\_13} (Compliance Officer)\quad
\textbf{Channel:} Financial Reporting Automation\\
\textbf{Phase:} Risk: Formatting Inconsistencies\quad
\textbf{Timestamp:} 2025-07-23T17:51:00

\smallskip
\begingroup\itshape
``Milestone check: we've officially detected the Formatting
Inconsistencies risk---our reports have now achieved the jazz hands
phase of compliance. At 52\% complete, the job is no longer just
mapping regulations; it's making sure headings, dates, and tables
stop playing costume changes between template, export, and
downstream views.

From my side, this is a useful turning point because consistency is
now a control issue, not just a presentation nuisance. I'll keep
pressure on standardizing the spec, and \textbf{I need Finance and
Data Engineering aligned on the same formatting rules} before this
turns into an audit-season sequel nobody asked for.''
\endgroup
\end{tcolorbox}

\paragraph{Baselines that answered correctly.}
The three correct answers come from very different ingestion pipelines.
What they share is that User\_13's request and named counterparties
remain reachable at retrieval time---either because nothing was
rewritten at all, or because the rewrite kept a speaker anchor.

\begin{tcolorbox}[
  title={\textbf{BM25 \textcolor{green!50!black}{$\checkmark$ Correct}}\quad\emph{(no semantic processing during ingestion; raw message indexed verbatim)}},
  colback=green!4, colframe=green!45!black,
  fonttitle=\bfseries,
  breakable,
]
\textbf{Top retrieved passage} (User\_13's actual message at rank 2):

\smallskip
\begin{quote}\ttfamily\small
[user=User\_13 / speaker\_role=Compliance Officer /
phase\_name=Risk: Formatting Inconsistencies /
timestamp=2025-07-23T17:51:00 / msg\_node=\textbf{Msg\_28294}]\\
\itshape
Milestone check\ldots\ \textbf{I need Finance and Data Engineering
aligned on the same formatting rules}\ldots
\end{quote}

\smallskip
\textbf{Agent answer:}\quad ``Finance and Data Engineering.''\\
\textbf{Why it works:} the answer literally appears in the retrieved
passage, with all metadata intact (\texttt{User\_13},
\texttt{Compliance Officer}, exact phase + timestamp).
\end{tcolorbox}

\begin{tcolorbox}[
  title={\textbf{text-embedding-3-large \textcolor{green!50!black}{$\checkmark$ Correct}}\quad\emph{(dense retrieval over raw messages)}},
  colback=green!4, colframe=green!45!black,
  fonttitle=\bfseries,
  breakable,
]
The dense retriever ranks several semantically similar messages from
\emph{other} users above the answer message. \texttt{Msg\_28294}
itself appears only at \textbf{rank 7} of the top-10:

\smallskip
\begin{quote}\ttfamily\small
[1] User\_7 / Data Analyst / Risk: Formatting Inconsistencies\\
{}[2] User\_7 / Data Analyst / Risk: Formatting Inconsistencies\\
{}[3] User\_13 / Compliance Officer / 2025-07-19 (\texttt{Msg\_1545})\\
\ldots\\
\textbf{[7] User\_13 / Compliance Officer / 2025-07-23 (\texttt{Msg\_28294}) \quad ←\quad answer here}
\end{quote}

\smallskip
\textbf{Agent answer:}\quad ``Finance and Data Engineering.''\\
\textbf{Why it works:} the gpt-5 agent reads the full top-10 context
and surfaces the correct phrasing from rank 7. The pipeline survives
because nothing was rewritten---it just relied on a longer effective
window than BM25 did.
\end{tcolorbox}

\begin{tcolorbox}[
  title={\textbf{hindsight \textcolor{green!50!black}{$\checkmark$ Correct}}\quad\emph{(LLM-rewritten ``experience'' that explicitly retains the speaker anchor)}},
  colback=green!4, colframe=green!45!black,
  fonttitle=\bfseries,
  breakable,
]
\textbf{Top retrieved memory:}

\smallskip
\begin{quote}\ttfamily\small
[type=experience / msg\_node=Msg\_28294 / mentioned\_at=2025-07-23T17:51 /
phase=Risk: Formatting Inconsistencies]\\
\itshape
``\textbf{User\_13 plans to standardize formatting specifications and
needs alignment from Finance and Data Engineering on formatting
rules} to avoid future audit issues. Involving: User\_13 (Compliance
Officer).''
\end{quote}

\smallskip
\textbf{Agent answer:}\quad ``Finance and Data Engineering.''\\
\textbf{Why it works:} hindsight's ingestion compresses the original
post but keeps the speaker (\texttt{User\_13}) bound to the request
and to the named counterparties. Compression is heavy, but the three
load-bearing fields---\emph{who}, \emph{what they need}, \emph{from
whom}---all survive.
\end{tcolorbox}

\paragraph{Baselines that answered incorrectly.}
The five incorrect answers fall into three distinct failure modes:
\textit{topic-stub compression} (mem0), \textit{wrong-speaker
shadowing} (a-mem), and \textit{multi-speaker over-broadening}
(hipporag, memgpt, graphrag).

\begin{tcolorbox}[
  title={\textbf{mem0 \textcolor{red!70!black}{$\times$ Incorrect}}\quad\emph{(LLM rewrites each message into an extracted ``memory''---failure mode: topic-stub compression)}},
  colback=red!4, colframe=red!50!black,
  fonttitle=\bfseries,
  breakable,
]
The original \texttt{Msg\_28294} is \emph{not} in the top retrievals.
Three unrelated messages have all been collapsed by the rewriter into
a near-identical topic stub:

\smallskip
\begin{quote}\ttfamily\small
[\textbf{User\_4} / IT Systems Lead / phase=Risk:
\textbf{Calculation Discrepancy}]\quad ``Phase is Risk: Formatting
Inconsistencies''\\
{}[\textbf{User\_7} / Data Analyst / phase=Risk:
\textbf{Regulatory Compliance Gap}]\quad ``Phase is Risk: Formatting
Inconsistencies''\\
{}[\textbf{User\_3} / Product Owner / channel=\textbf{Sustainable Finance
Strategy}]\quad ``Requesting senior alignment on final wording \ldots''
\end{quote}

\smallskip
\textbf{What was lost.}
\begin{itemize}\setlength\itemsep{0pt}
\item The \emph{speaker} (\texttt{User\_13}) is no longer attached to
      the request.
\item The \emph{request itself} (``I need \ldots aligned'') is
      compressed away; only the topic label survives.
\item The \emph{counterparty} (Finance and Data Engineering) is gone.
\end{itemize}

\smallskip
\textbf{Agent answer:}\quad ``Legal, Risk, and Reporting teams.''\\
The agent fills the gap with a generic guess from compliance-officer
priors because the retrieved passages no longer mention any specific
counterparty.
\end{tcolorbox}

\begin{tcolorbox}[
  title={\textbf{a-mem \textcolor{red!70!black}{$\times$ Incorrect}}\quad\emph{(failure mode: wrong-speaker shadowing)}},
  colback=red!4, colframe=red!50!black,
  fonttitle=\bfseries,
  breakable,
]
All three top-ranked memories are \texttt{User\_7}'s posts on the same
topic; the asker's own message never surfaces:

\smallskip
\begin{quote}\ttfamily\small
[1] Author: \textbf{User\_7} / Data Analyst / phase=Risk: Formatting Inconsistencies (early)\\
{}[2] Author: \textbf{User\_7} / Data Analyst / phase=Risk: Formatting Inconsistencies (mid)\\
{}[3] Author: \textbf{User\_7} / Data Analyst / phase=Risk: Formatting Inconsistencies (late)
\end{quote}

\smallskip
\textbf{What was lost.} Speaker identity isn't physically erased
(\texttt{Author: User\_7} is in every retrieved memory) but it
\emph{has been ignored at retrieval time}: similarity-search returned
three near-duplicate posts about the same topic from a single louder
speaker, and shadowed User\_13's actual request entirely.

\smallskip
\textbf{Agent answer:}\quad (empty)\\
The agent declines to answer because none of the retrieved User\_7
posts name a counterparty for User\_13.
\end{tcolorbox}

\begin{tcolorbox}[
  title={\textbf{hipporag \textcolor{red!70!black}{$\times$ Incorrect}}\quad\emph{(failure mode: multi-speaker over-broadening, KG-backed)}},
  colback=red!4, colframe=red!50!black,
  fonttitle=\bfseries,
  breakable,
]
The personalized-PageRank traversal returns three different speakers'
posts about the same risk:

\smallskip
\begin{quote}\ttfamily\small
[1] \textbf{User\_12} (Compliance Officer): ``\ldots Please weigh in
from \textbf{Finance and Data Engineering} \ldots''\\
{}[2] \textbf{User\_4} (IT Systems Lead): ``\ldots cross-functional
review: \textbf{Finance, Data Engineering, QA, and template owners}\ldots''\\
{}[3] \textbf{User\_12} (Compliance Officer): ``\ldots I need
\textbf{Finance and Engineering} to confirm \ldots''
\end{quote}

\smallskip
\textbf{What was lost.} The right entities (Finance, Data Engineering)
\emph{are} in the retrieval, but they are scattered across three
\emph{other} users' requests, each with slightly different
counterparty lists. The agent unions the candidate sets rather than
honoring the asker's specific request.

\smallskip
\textbf{Agent answer:}\quad ``Finance, Data Engineering, QA, and the
report/template owners (with Compliance for sign-off).''\\
The judge marks this incorrect because the union goes beyond the gold
answer.
\end{tcolorbox}

\begin{tcolorbox}[
  title={\textbf{memgpt \textcolor{red!70!black}{$\times$ Incorrect}}\quad\emph{(failure mode: multi-speaker over-broadening, archival-passage retrieval)}},
  colback=red!4, colframe=red!50!black,
  fonttitle=\bfseries,
  breakable,
]
Like a-mem, memgpt's archival retrieval returns three User\_7 / User\_4
messages on the topic. Unlike a-mem, the agent extracts every
counterparty term it sees and concatenates them all:

\smallskip
\begin{quote}\ttfamily\small
[1] \textbf{User\_7} (Data Analyst, early phase)\\
{}[2] \textbf{User\_7} (Data Analyst, late phase)\\
{}[3] \textbf{User\_4} (IT Systems Lead, late phase)
\end{quote}

\smallskip
\textbf{Agent answer:}\quad ``\textbf{Finance}, Operations, Reporting
Owners, Compliance, \textbf{Data Engineering}, QA, and Template
Owners.''\\
\textbf{Why it fails.} Same root cause as hipporag: the asker's
specific request was never retrieved, so the agent assembled a
``who-has-ever-been-mentioned'' list. The two correct names
(\textbf{Finance}, \textbf{Data Engineering}) are in there, but they
are diluted by five extras the gold answer doesn't include.
\end{tcolorbox}

\begin{tcolorbox}[
  title={\textbf{graphrag \textcolor{red!70!black}{$\times$ Incorrect}}\quad\emph{(failure mode: community-summary abstraction)}},
  colback=red!4, colframe=red!50!black,
  fonttitle=\bfseries,
  breakable,
]
graphrag's local search returns a community summary rather than any
individual chat message:

\smallskip
\begin{quote}\ttfamily\small
[graphrag local context (per-domain index)]\\
-----Reports-----\\
\textbf{Treasury Management and Compliance Community} \ldots\
``centered around the critical roles of several key entities involved
in financial \ldots''
\end{quote}

\smallskip
\textbf{What was lost.} An entire layer of granularity. The community
report aggregates dozens of messages (from many speakers) into a
domain-level abstract; the asker, the specific request, and the named
counterparties have all been folded into a topical narrative.

\smallskip
\textbf{Agent answer:}\quad ``\textbf{Finance} (including Finance Ops),
QA, and \textbf{Data Engineering}/Engineering.''\\
The agent recovers the right entities but again over-broadens to a
generic compliance-counterparty list.
\end{tcolorbox}

\paragraph{Summary.}
This single question is enough to read off the design space: the
\emph{outcome} (correct or not) is set at \emph{ingestion time},
before retrieval or QA gets a chance to fix anything. The two
ingestion strategies that preserve the answer are, on the surface,
opposites---BM25 keeps every original token and metadata field;
hindsight rewrites aggressively but explicitly conditions every
``experience'' on its speaker. What unites them is that
\emph{User\_13's request and its named counterparty stay
co-located} in some retrievable unit. Each of the five failures
breaks that co-location in a different way: by collapsing the
content to a topic label (mem0), by ranking similar-topic posts
from a louder speaker above the asker's own (a-mem), or by
mixing several speakers' counterparty lists into a single
over-broad answer (hipporag, memgpt, graphrag). Section~\ref{sec:results}
quantifies the same pattern in aggregate; this appendix shows
its mechanics on a single instance.

\section{Limitations}
\label{appendix:limitations}
GroupMemBench currently focuses on English-language, text-only workplace conversations across four domains (Technology, Healthcare, Manufacturing, Finance). Two natural extensions are left to future work: a \emph{multilingual} setting, where speaker-conditioned vocabulary and Theory-of-Mind effects interact with code-switching and translation drift; and a \emph{multimodal} setting, where group chats also contain images, attachments, and voice notes, requiring memory systems to ground references across modalities. We expect both extensions to be additive on top of the graph-grounded synthesis pipeline introduced here, and the design principles of speaker-aware ingestion, audience-adapted language, and reply-structure control transfer directly to those settings.

\section{Broader Impacts}
\label{appendix:broader_impacts}
GroupMemBench is intended to drive research on agent memory in realistic multi-party settings, with potential positive impact on team collaboration tools such as channel assistants, project copilots, and meeting summarizers, where better speaker-grounded memory can reduce miscommunication and lower the cognitive load of long-running discussions. On the negative side, stronger speaker-conditioned memory could in principle support workplace surveillance or more targeted persuasion if irresponsibly deployed; we mitigate this at the artifact level by releasing only synthetic conversations grounded in fictional personas (Section~\ref{sec:dataset_construction}) rather than any real chat logs, and we disclose the full persona vocabularies and prompt templates (Appendices~\ref{appendix:graph_schema},~\ref{appendix:prompt}) so downstream users can audit and reweight before reuse.


\end{document}